\documentclass[11pt]{article}
\usepackage{amsfonts, amssymb, amsmath}
\usepackage{algorithm, algorithmic, amsthm}
\usepackage{graphicx}
\usepackage{color}
\usepackage{hyperref}
\usepackage{url}
\usepackage{graphicx, wrapfig}
\usepackage{booktabs}
\usepackage{caption}
\usepackage{multirow}
\usepackage[top=2.5cm, bottom=3cm, left=3cm, right=3cm]{geometry}

\usepackage{subfigure}

\newcommand{\q}{\ensuremath{\mathbf{q}}}

{%
\begin{list}{#1}{
\vspace{-\topsep}
\vspace{-\partopsep}
\setlength{\itemindent}{0cm}
\setlength{\rightmargin}{0cm}
\setlength{\listparindent}{0cm}
\settowidth{\labelwidth}{#1}
\setlength{\leftmargin}{\labelwidth}
\addtolength{\leftmargin}{\labelsep}
\setlength{\itemsep}{0cm}
}%
}%
{%
\end{list}
\vspace{-\topsep}
\vspace{-\partopsep}
}

{\begin{enumerate}%
}%
{\end{enumerate}}

\theoremstyle{plain}

\newtheorem*{lemma*}{Lemma}

\newtheorem*{prop*}{Proposition}

\theoremstyle{definition}

\newtheorem*{defn*}{Definition}

\newtheorem*{exmp*}{Example}

\newtheorem*{conj*}{Conjecture}

\theoremstyle{remark}

\newtheorem*{rmk*}{Remark}

\def\model/{{\sc Neural Enquirer}}
\def\modelm/{{\sc Neural Enquirer}-{M}}
\def\q/{\mathbf{q}}
\def\fname/{\mathbf{f}}
\def\sempre/{{\sc Sempre}}

\def\selwh/{{\sc Select\_Where}}
\def\super/{{\sc Superlative}}
\def\whsuper/{{\sc Where\_Superlative}}
\def\nested/{{\sc Nest}}

\title{{\sc Neural Enquirer}: Learning to Query Tables with Natural Language}

\author{\sf Pengcheng Yin$^\dagger$\thanks{Work done when the first author worked as an intern at Noah's Ark Lab, Huawei Technologies.} \ \;Zhengdong Lu$^\ddagger$\; Hang Li$^\ddagger$ \; Ben Kao$^\dagger$ \\ $^\dagger$Dept. of Computer Science\\ The University of Hong Kong\\ {\tt \{pcyin, kao\}@cs.hku.hk}\\ $^\ddagger$Noah's Ark Lab, Huawei Technologies\\ {\tt \{Lu.Zhengdong, HangLi.HL\}@huawei.com} }

\date{}
\begin{document}

\maketitle

\begin{abstract}


We proposed \model/ as a neural network architecture to execute a natural language (NL) query on a knowledge-base (KB) for answers. Basically, \model/ finds the distributed representation of a query and then executes it on knowledge-base tables to obtain the answer as one of the values in the tables. Unlike similar efforts in end-to-end training of semantic parsers~\cite{pasupat2015compositional,2015arXiv151104834N}, \model/ is fully ``neuralized": it not only gives distributional representation of the query and the knowledge-base, but also realizes the execution of compositional queries as a series of differentiable operations, with intermediate results (consisting of annotations of the tables at different levels) saved on multiple layers of memory. \model/ can be trained with gradient descent, with which not only the parameters of the controlling components and semantic parsing component, but also the embeddings of the tables and query words can be learned from scratch. The training can be done in an end-to-end fashion, but it can take stronger guidance, e.g., the step-by-step supervision for complicated queries, and benefit from it. \model/ is one step towards building neural network systems which seek to understand language by executing it on real-world. Our experiments show that \model/ can learn to execute fairly complicated NL queries on tables with rich structures.

\end{abstract}


\section{Introduction}

In models for natural language dialogue and question answering, there is ubiquitous need for querying a knowledge-base~\cite{DBLP:conf/emnlp/WenGMSVY15,pasupat2015compositional}. The traditional pipeline is to put the query through a semantic parser to obtain some ``executable" representations, typically logical forms, and then apply this representation to a knowledge-base for the answer. Both the semantic parsing and the query execution part can get quite messy for complicated queries like $\tilde{Q}$: ``\textit{Which city hosted the longest game before the game in Beijing?}" in Figure~\ref{fig:model_overview}, and need carefully devised systems with hand-crafted features or rules to derive the correct logical form $\tilde{F}$ (written in SQL-like style). Partially to overcome this difficulty, there has been effort~\cite{pasupat2015compositional} to ``backpropagate" the result of query execution to revise the semantic representation of the query, which actually falls into the thread of work on learning from grounding~\cite{chen:icml08}. One drawback of these semantic parsing models is rather symbolic with rule-based features, leaving only a handful of tunable parameters to cater to the supervision signal from the execution result.

On the other hand, neural network-based models are previously successful mostly on tasks with direct and strong supervision in natural language processing or related domain, with examples including machine translation and syntactic parsing. The recent work on learning to execute simple Python code with LSTM~\cite{DBLP:journals/corr/ZarembaS14} pioneers in the direction on learning to parse structured objects through executing it in a purely neural way, while the later work on Neural Turing Machine (NTM)~\cite{DBLP:journals/corr/GravesWD14} introduces more modeling flexibility by equipping the LSTM with external memory and various means of interacting with it.

Our work, inspired by above-mentioned threads of research, aims to design a neural network system that can learn to understand the query and execute it on a knowledge-base table from examples of queries and answers.
Our proposed \model/ encodes queries and KBs into distributed representations, and executes compositional queries against the KB through a series of differentiable operations. It can be trained using Query-Answer pairs, where the distributed representations of queries and the KB are optimized together with the query execution logic in an end-to-end fashion.
We then demonstrates using a synthetic question-answering task that our proposed \model/ is capable of learning to execute compositional natural language queries with complex structures.


\section{Overview of Neural Enquirer}

\label{overview}

Given an NL query $Q$ and a KB table $\mathcal{T}$, \model/ executes the query against the table and outputs a ranked list of query answers.
The execution is done by first using {\em Encoders} to encode the query and table into distributed representations, which are then sent to a cascaded pipeline of {\em Executors} to derive the answer.
Figure \ref{fig:model_overview} gives an illustrative example (with five executors) of various types of components involved:

\begin{figure*}[tb]
	\centering
	\includegraphics[width=\textwidth]{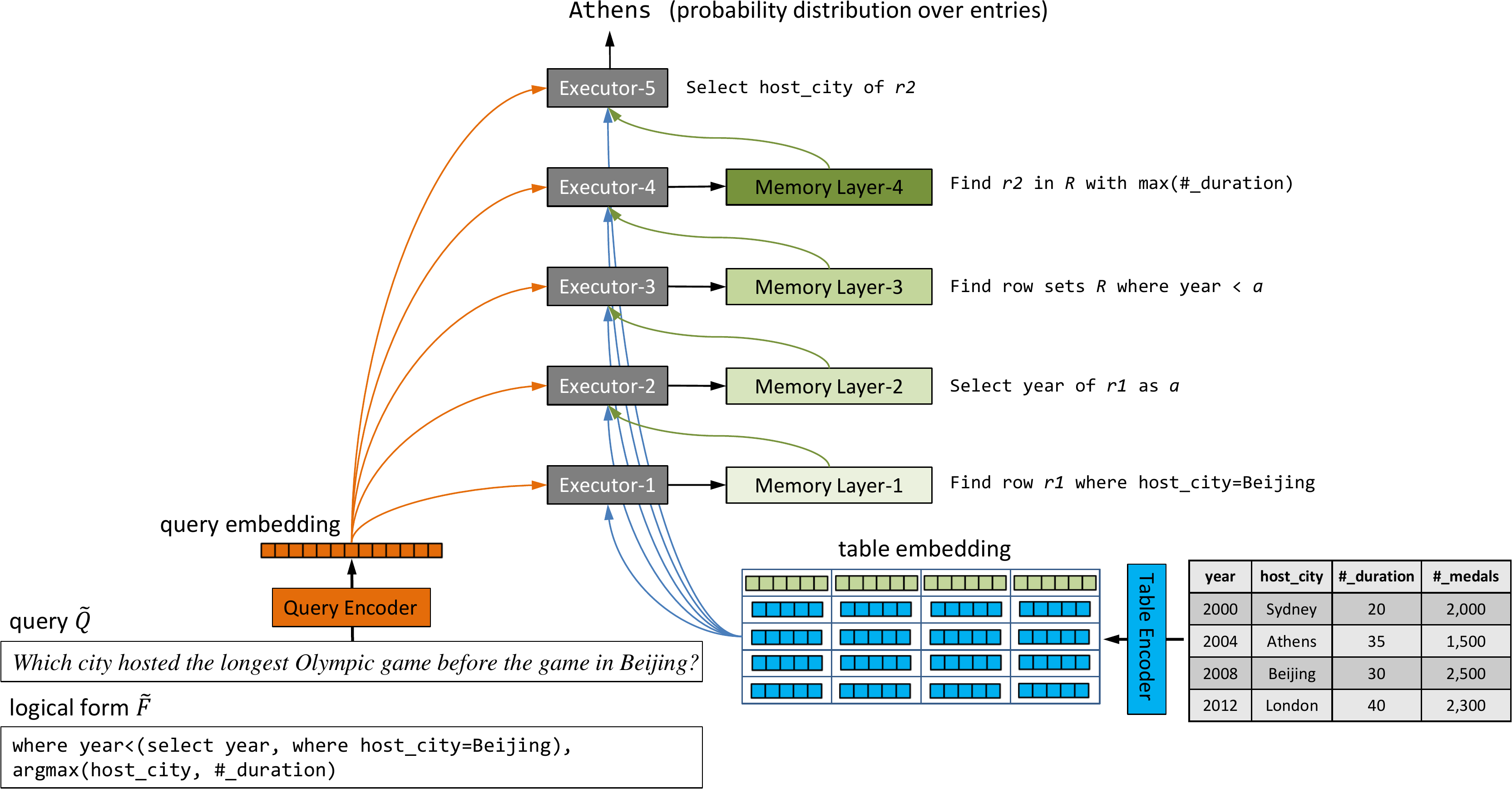}
	\caption{An overview of \model/ with five executors}
	\label{fig:model_overview}
\end{figure*}

\phantom{xx}
\noindent {\bf Query Encoder} (Section \ref{query_encoder}), which encodes the query into a distributed representation that carries the semantic information of the original query. The encoded query embedding will be sent to various executors to compute its execution result.

\phantom{xx}
\noindent {\bf Table Encoder} (Section \ref{table_encoder}), which encodes entries in the table into distributed vectors. Table Encoder outputs an embedding vector for each table entry, which retains the two-dimensional structure of the table.

\phantom{xx}
\noindent {\bf Executor} (Section \ref{executor}), which executes the query against the table and outputs {\em annotations} that encode intermediate execution results, which are stored in the memory of each layer to be accessed by subsequent executor. 
Our basic assumption is that complex, compositional queries can be answered through multiple steps of computation, where each executor models a specific type of operation conditioned on the query. Figure~\ref{fig:model_overview} illustrates the operation each executor is supposed to perform in answering $\tilde{Q}$. Different from classical semantic parsing approaches which require a predefined set of all possible logical operations, \model/ is capable of {\em learning} the logic of executors via end-to-end training using Query-Answer pairs.
By stacking several executors, \model/ is able to answer complex queries involving multiple steps of computation.


\section{Model}
In this section we give a more detailed exposition of different types of components in the \model/ model.


\subsection{Query Encoder}
\label{query_encoder}

Given an NL query $Q$ composed of a sequence of words $\{ w_1, w_2, \ldots, w_T \}$, Query Encoder parses $Q$ into a $d_\mathcal{Q}$-dimensional vectorial representation $\q/$: $Q\xrightarrow{\tt encode}\q/ \in \mathbb{R}^{d_\mathcal{Q}}$.
In our implementation of \model/, we employ a bidirectional RNN for this mission\footnote{Other choices of sentence encoder such as LSTM or even convolutional neural networks are possible too}. 
More specifically, the RNN summarizes the sequence of word embeddings of $Q$, $\{ \mathbf{x}_1,\mathbf{x}_2, \ldots, \mathbf{x}_T \}$, into a vector $\q/$ as the representation of $Q$, where $\mathbf{x}_t = \mathbf{L}[w_t]$, $\mathbf{x}_t \in \mathbb{R}^{d_{\mathcal{W}}}$ and $\mathbf{L}$ is the embedding matrix. See Appendix~\ref{app:gru} for details.

It is worth noting that our Query Encoder can find the representation of rather general class of symbol sequences, agnostic to the actual representation of queries (e.g., natural language, SQL-like, etc). \model/ is capable of learning the execution logic expressed in the input query through end-to-end training, making it a generic model for query execution.

\subsection{Table Encoder}
\label{table_encoder}

\begin{wrapfigure}{r}{0.25\textwidth}
	\vspace{-25pt}
    \includegraphics[scale=0.65]{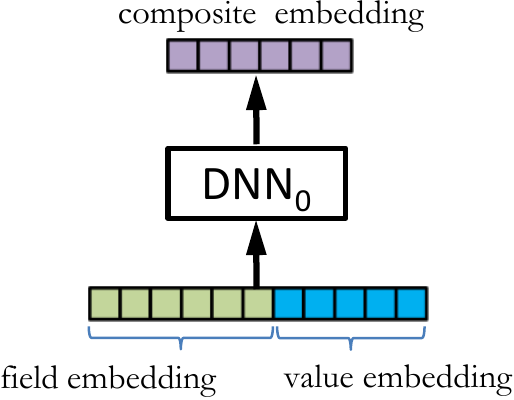}
    \vspace{-15pt}
\end{wrapfigure}
Table Encoder converts a knowledge-base table $\mathcal{T}$ into its distributional representation as  input to \model/. Suppose the table has $M$ rows and $N$ columns, where each column comes with a {\em field name} (e.g., {\tt host\_city}), and the value of each table entry is a word (e.g., {\tt Beijing}) in our vocabulary,
 Table Encoder first finds the embedding for field names and values of table, and then it computes the (field, value) composite embedding for each of the $M\times N$ entries in the table. More specifically, for the entry in the $m$-th row and $n$-th column with a value of $w_{mn}$, Table Encoder computes a $d_{\mathcal{E}}$-dimensional embedding vector $\mathbf{e}_{mn}$ by fusing the embedding of the entry value with the embedding of its corresponding field name as follows:
\[ \mathbf{e}_{mn} = {\sf DNN}_0([\mathbf{L}[w_{mn}];\fname/_n]) = \tanh (\mathbf{W}_{\it} \cdot [\mathbf{L}[w_{mn}]; \fname/_n] + \mathbf{b}) \]
where $\fname/_n$ is the embedding of the field name (of the $n$-th column). $\mathbf{W}$ and $\mathbf{b}$ denote the weight matrices, and $[\cdot ; \cdot]$ the concatenation of vectors. The output of Table Encoder is a tensor of shape $M \times N \times d_{\mathcal{E}}$, consisting of $M \times N$ embeddings of length $d_{\mathcal{E}}$ for all entries.

Our Table Encoder functions differently from classical knowledge embedding models (e.g., {\sf TransE}~\cite{DBLP:conf/nips/BordesUGWY13}), where embeddings of entities (entry values) and relations (field names) are learned in a unsupervised fashion via minimizing certain reconstruction errors. Embeddings in \model/ are optimized via supervised learning towards end-to-end QA tasks. Additionally, as will shown in the experiments, those embeddings function in a way as indices, which not necessarily encode the exact semantic meaning of their corresponding words.


\subsection{Executor}
\label{executor}

\begin{figure}[tb]
	\centering
	\includegraphics[scale=0.6]{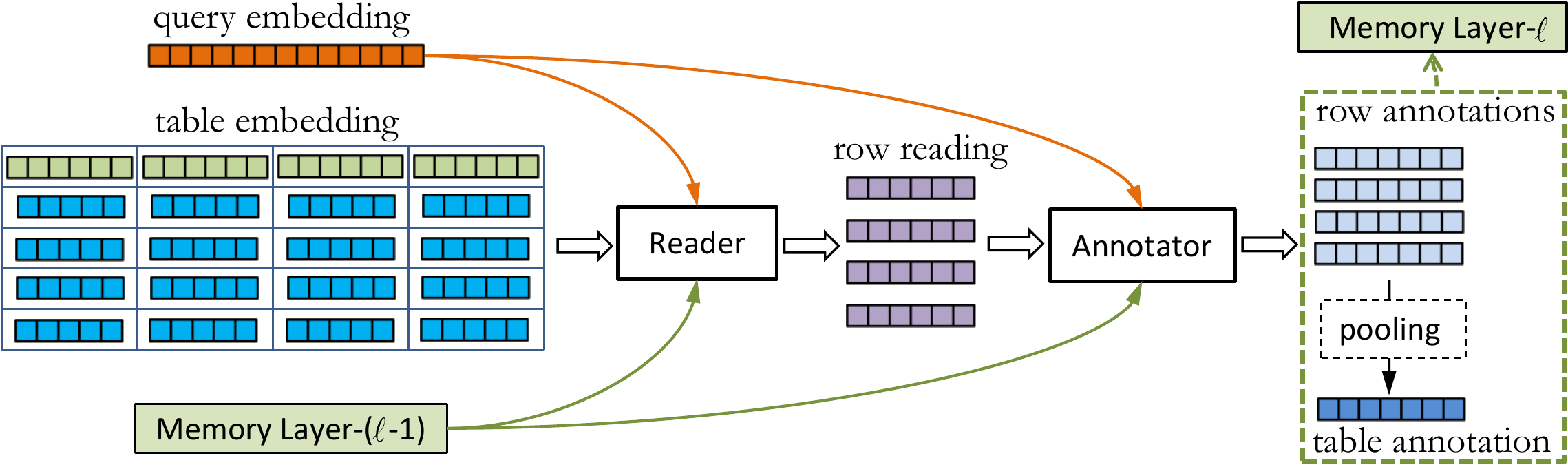}
	\caption{Overview of an {\sf Executor-$\ell$}}
	\label{fig:executor}
\end{figure}

\model/ executes an input query on a KB table through layers of execution. Each layer of executor captures a certain type of operation (e.g., {\tt select}, {\tt where}, {\tt max}, etc.) relevant to the input query\footnote{Depending on the query, an executor may perform different operations.}, and returns intermediate execution results, referred to as {\em annotations}, saved in an external memory of the same layer. 
A query is executed step-by-step through a sequence of stacked executors. Such a cascaded architecture enables \model/ to answer complex, compositional queries.
An illustrative example is given in Figure~\ref{fig:model_overview}, with each executor annotated with the operation it is assumed to perform.
We will demonstrate in Section~\ref{sec:exp} that \model/ is capable of learning the operation logic of each executor via end-to-end training.


As illustrated in Figure~\ref{fig:executor}, an executor at Layer-$\ell$ (denoted as {\sf Executor-$\ell$})
has two major neural network components: Reader and Annotator. An executor processes a table row-by-row.
For the $m$-th row, with $N$ (field, value) composite embeddings $\mathcal{R}_m=\{\mathbf{e}_{m1}, \mathbf{e}_{m2}, \ldots, \mathbf{e}_{mN}\}$, 
the Reader fetches a read vector $\mathbf{r}^{\ell}_{m}$ from $\mathcal{R}_m$, 
which is sent to the Annotator to compute a \emph{row annotation} $\mathbf{a}^{\ell}_{m} \in \mathbb{R}^{d_{\mathcal{A}}}$:
\begin{eqnarray}
  \textsf{Read Vector:} &&\mathbf{r}^{\ell}_{m} = f^{\ell}_\textsc{r}(\mathcal{R}_m, \mathcal{F}_\mathcal{T}, \q/, \mathcal{M}^{\ell-1})\\
  \textsf{Row Annotation:} &&\mathbf{a}^{\ell}_{m} = f^{\ell}_\textsc{a}(\mathbf{r}^{\ell}_{m}, \q/, \mathcal{M}^{\ell-1})
\end{eqnarray}
where $\mathcal{M}^{\ell-1}$ denotes the content in memory Layer-$(\ell\hspace{-3pt}-\hspace{-3pt}1)$, and $\mathcal{F}_\mathcal{T}= \{\fname/_1, \fname/_2, \ldots, \fname/_N\}$ is the set of field name embeddings.  Once all row annotations are obtained, {\sf Executor-$\ell$} then generates the \emph{table annotation} through the following pooling process:
\[
  \textsf{Table Annotation:} ~~~ \mathbf{g}^{\ell} = f_{\textsc{Pool}}(\mathbf{a}^{\ell}_1, \mathbf{a}^{\ell}_2, \ldots, \mathbf{a}^{\ell}_M).
\]
A row annotation captures the local execution result on each row, while a table annotation, derived from all row annotations, summarizes the global computational result on the whole table. Both row annotations $\{\mathbf{a}^{\ell}_1, \mathbf{a}^{\ell}_2, \ldots, \mathbf{a}^{\ell}_M\}$ and table annotation $\mathbf{g}^{\ell} $ are saved in memory Layer-$\ell$: $\mathcal{M}^{\ell} = \{\mathbf{a}^{\ell}_1, \mathbf{a}^{\ell}_2, \ldots, \mathbf{a}^{\ell}_M, \mathbf{g}^{\ell} \}$.

\begin{figure}[tb]
	\centering
	\includegraphics[scale=0.6]{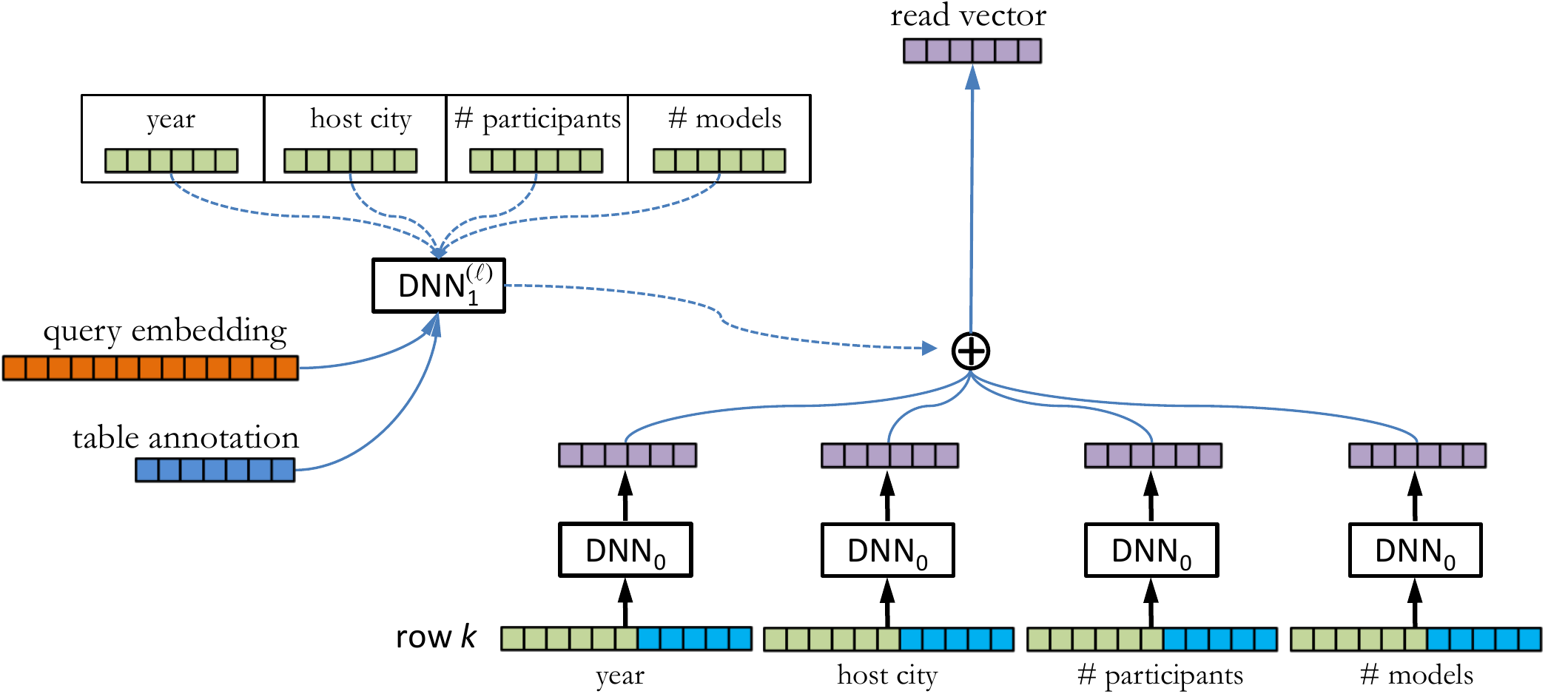}
	\caption{Illustration of the Reader for \textsf{Executor}-$\ell$.}
	\label{fig:content_based_read}
\end{figure}

Our design of executor is inspired by Neural Turing Machines~\cite{DBLP:journals/corr/GravesWD14}, where data is fetched from an external memory using a read head, and subsequently processed by a controller, whose outputs are flushed back in to memories.
An executor functions similarly by reading data from each row of the table, using a Reader, and then calling an Annotator to calculate intermediate computational results as annotations, which are stored in the executor's memory. We assume that row annotations are able to handle operations which require only row-wise, local information (e.g., {\tt select}, {\tt where}), while table annotations can model superlative operations (e.g., {\tt max}, {\tt min}) by aggregating table-wise, global execution results. Therefore, a combination of row and table annotations enables \model/ to capture a variety of real-world query operations.

\subsubsection{Reader}
As illustrated in Figure~\ref{fig:content_based_read}, an executor at Layer-$\ell$ reads in a vector $\mathbf{r}_m^\ell$ for each row $m$, defined as the weighted sum of composite embeddings for entries in this row:
\[ \mathbf{r}^{\ell}_{m} = f^{\ell}_\textsc{r}(\mathcal{R}_m, \mathcal{F}_\mathcal{T}, \q/, \mathcal{M}^{\ell-1}) = \sum^N_{n=1}\tilde{\omega}(\fname/_n, \q/, \mathbf{g}^{\ell-1})\mathbf{e}_{mn} \]
where $\tilde{\omega}(\cdot)$ is the normalized attention weights given by:
\begin{equation}
  \label{eq:read_attention_weight}
  \tilde{\omega}(\fname/_n, \q/, \mathbf{g}^{\ell-1})=\frac{\exp( \omega(\fname/_n, \q/, \mathbf{g}^{\ell-1}) )}{\sum^N_{n'=1}\exp( \omega(\fname/_{n'}, \q/, \mathbf{g}^{\ell-1}) )}\
\end{equation}
and $\omega(\cdot)$ is modeled as a DNN (denoted as ${\sf DNN}_1^{(\ell)}$).

Note that the $\tilde{\omega}(\cdot)$ is agnostic to the values of entries in the row, i.e., in an executor all rows share the same set of weights $\tilde{\omega}(\cdot)$. Since each executor models a specific type of computation, it should only attend to a subset of entries pertain to its execution, which is modeled by the Reader. This is related to the content-based addressing of Neural Turing Machines~\cite{DBLP:journals/corr/GravesWD14} and the attention mechanism in neural machine translation models~\cite{chonmt}.

\subsubsection{Annotator} In \textsf{Executor}-$\ell$, the Annotator computes row and table annotations based on the fetched read vector $\mathbf{r}^{\ell}_{m}$ of the Reader, which are then stored in the $\ell$-th memory layer $\mathcal{M}^\ell$ accessible to {\sf Executor-$(\ell\hspace{-3pt}+\hspace{-3pt}1)$}. 
This process is repeated in intermediate layers, until the executor in the last layer to finally generate the answer.

\begin{figure}[h!]
	\centering
	\includegraphics[scale=0.6]{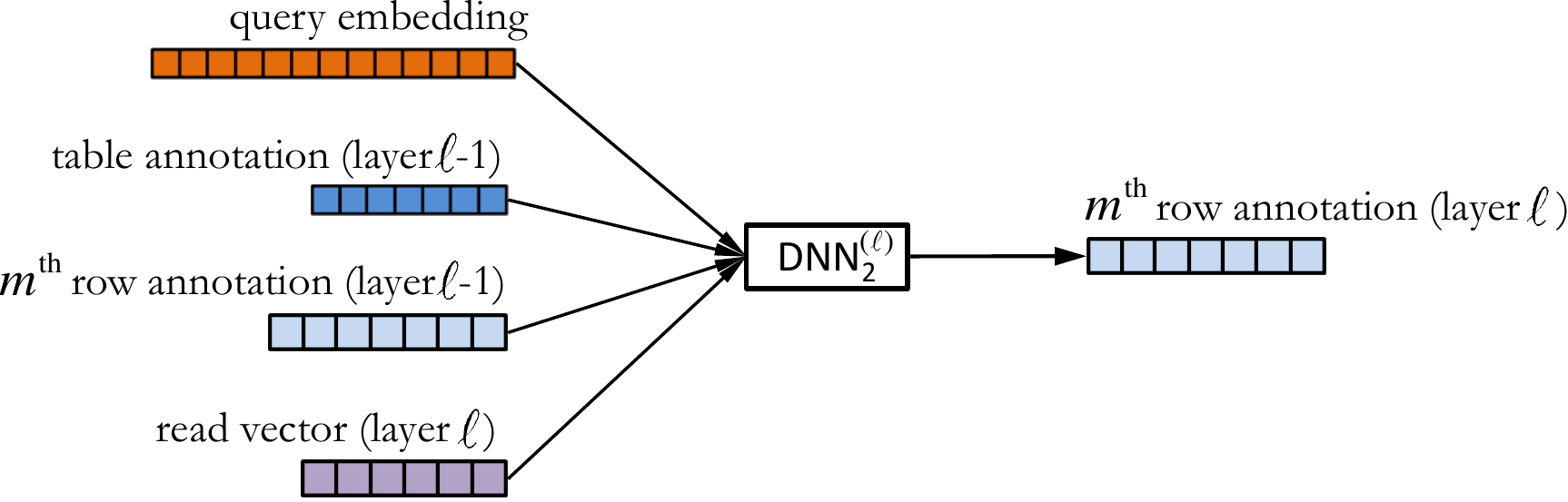}
	\caption{Illustration of Annotator for \textsf{Executor}-$\ell$.}
	\label{fig:row_annotation_computation}
\end{figure}

\paragraph{{Row annotations}}
A row annotation encodes the local computational result on a specific row. As illustrated in Figure~\ref{fig:row_annotation_computation}, a row annotation for row $m$ in \textsf{Executor}-$\ell$, given by 
\begin{equation}
  \label{eq:row_annot}
  \mathbf{a}^{\ell}_m = f^{\ell}_\textsc{a}(\mathbf{r}^{\ell}_{m}, \q/, \mathcal{M}^{\ell-1}) = \textsf{DNN}_2^{(\ell)}([\mathbf{r}^{\ell}_{m}; \q/; \mathbf{a}^{\ell-1}_m; \mathbf{g}^{\ell-1}]).
\end{equation}
fuses the corresponding read vector $\mathbf{r}^{\ell}_{m}$, the results saved in previous memory layer (row and table annotations $\mathbf{a}_m^{\ell-1}$, $\mathbf{g}^{\ell-1}$), and the query embedding $\q/$. Basically,  
\begin{itemize}
  \item row annotation $\mathbf{a}_m^{\ell-1}$ represents the local status of the execution before Layer-$\ell$;
  \item table annotation $\mathbf{g}^{\ell-1}$ summarizes the global status of the execution before Layer-$\ell$;
  \item read vector $\mathbf{r}^{\ell}_{m}$ stores the value of current attention;
  \item query embedding $\q/$ encodes the overall execution agenda,
\end{itemize}
all of which are combined through $\textsf{DNN}_2^{(\ell)}$ to form the annotation of row $m$ in the current layer.

\paragraph{{Table annotations}} Capturing the global execution state, a table annotation is summarized from all row annotations via a global pooling operation. In our implementation of \model/ we employ max pooling:
\begin{equation}
  \label{eq:pooling}
  \mathbf{g}^{\ell} = f_{\textsc{Pool}}(\mathbf{a}^{\ell}_1, \mathbf{a}^{\ell}_2, \ldots, \mathbf{a}^{\ell}_M) = [g_1, g_2, \ldots, g_{d_{\mathcal{G}}}]^\top
\end{equation}
where $ g_k = \max(\{ \mathbf{a}^{\ell}_1(k), \mathbf{a}^{\ell}_2(k), \ldots, \mathbf{a}^{\ell}_M(k) \})$ is the maximum value among the $k$-th elements of all row annotations. It is possible to use other pooling operations (e.g., gated pooling), but we find max pooling yields the best results.

\subsubsection{Last Layer of Executor} Instead of computing annotations based on read vectors, the last executor in \model/ directly outputs the probability of the value of each entry in $\mathcal{T}$ being the answer:
\begin{equation}
  \label{eq:answer_prob}
  p(w_{mn}|Q, \mathcal{T}) = \frac{\exp (f_\textsc{Ans}^{\ell}(\mathbf{e}_{mn}, \q/, \mathbf{a}^{\ell - 1}_m, \mathbf{g}^{\ell - 1}))}{ \sum_{m'=1}^M \sum_{n'=1}^N \exp (f_\textsc{Ans}^{\ell}(\mathbf{e}_{m'n'}, \q/, \mathbf{a}^{\ell - 1}_{m'}, \mathbf{g}^{\ell - 1})) }
\end{equation}
where $f_\textsc{ans}^{\ell}(\cdot)$ is modeled as a DNN.
Note that the last executor, which is devoted to returning answers, carries out a specific kind of execution using $f^\ell_\textsc{Ans} (\cdot)$ based on the entry value, the query, and annotations from previous layer. 


\subsection{Handling Multiple Tables}
Real-world KBs are often modeled by a schema involving various tables, where each table stores a specific type of factual information.
We present \modelm/, adapted for simultaneously operating on multiple KB tables. 
A key challenge in this scenario is that the multiplicity of tables requires modeling interaction between them. For example, \modelm/ needs to serve {\em join} queries, whose answer is derived by joining fields in different tables.
Details of the modeling and experiments of \modelm/ are given in Appendix~\ref{app:multi_table}.



\section{Learning}

\model/ can be trained in an {\it end-to-end} (N2N) fashion in Question Answering tasks. During training, both the representations of queries and table entries, as well as the execution logic captured by weights of executors are learned. More specifically, given a set of $N_{\mathcal{D}}$ query-table-answer triples $\mathcal{D} = \{ (Q^{(k)}, \mathcal{T}^{(i)}, y^{(i)}) \}$, we optimize the model parameters by maximizing the log-likelihood of gold-standard answers:
\begin{equation}
  \label{eq:ete_objective}
  \mathcal{L}_\text{N2N}(\mathcal{D}) = \sum_{i=1}^{N_\mathcal{D}} \log p(y^{(i)}=w_{mn}|Q^{(i)}, \mathcal{T}^{(i)})
\end{equation}
In end-to-end training, each executor discovers its operation logic from training data in a purely data-driven manner, which could be difficult for complicated queries requiring four or five sequential operations.

This can be alleviated by softly guiding the learning process via controlling the attention weights $\tilde{w}(\cdot)$ in Eq.~\eqref{eq:read_attention_weight}. By enforcing $\tilde{w}(\cdot)$ to bias towards a field pertain to a specific operation, we can ``coerce'' the executor to figure out the logic of this particular operation relative to the field.
As an example, for {\sf Executor}-1 in Figure \ref{fig:model_overview}, by biasing the weight of the {\tt host\_city} field towards 1.0, only the value of {\tt host\_city} field will be fetched and sent for computing annotations, in this way we can force the executor to learn to find the row whose {\tt host\_city} is {\it Beijing}. 
This setting will be referred to as {\it step-by-step} (SbS) training. Formally,
this is done by introducing additional supervision signal to Eq. \eqref{eq:ete_objective}:
\begin{equation}
  \label{eq:sbs_objective}
  \mathcal{L}_{\text{SbS}}(\mathcal{D}) = \sum_{i=1}^{N_\mathcal{D}}[ \log p(y^{(i)}=w_{mn}|Q^{(i)}, \mathcal{T}^{(i)}) +
  \alpha \sum_{\ell = 1}^L \log \tilde{w}(\mathbf{f}_{k, \ell}^{\star}, \cdot, \cdot) ]
\end{equation}
where $\alpha$ is a scalar and $\mathbf{f}_{k, \ell}^{\star}$ is the embedding of the field name known {\it a priori} to be relevant to the executor at Layer-$\ell$ in the $k$-th example. 


\section{Experiments}
\label{sec:exp}

In this section we evaluate \model/ on synthetic QA tasks with queries with varying compositional depth.
We will first briefly describe our synthetic QA task for benchmark and experimental setup, and then discuss the results under different settings.

\subsection{Synthetic QA Task}
We present a synthetic QA task to evaluate the performance of \model/, where a large amount of QA examples at various levels of complexity are generated to evaluate the single table and multiple tables cases of the model.
Starting with ``artificial'' tasks eases the process of developing novel deep models~\cite{DBLP:journals/corr/WestonBCM15}, and has gained increasing popularity in recent advances of the research on modeling symbolic computation using DNNs~\cite{DBLP:journals/corr/GravesWD14,DBLP:journals/corr/ZarembaS14}.

Our synthetic dataset consists of query-table-answer triples $\{(Q^{(i)}, \mathcal{T}^{(i)}, y^{(i)})\}$.
To generate such a triple, we first randomly sample a table $\mathcal{T}^{(i)}$ of size $10 \times 10$ from a synthetic schema of Olympic Games, which has 10 fields,
whose values are drawn from a vocabulary of size 240, with 120 country and city names, and 120 numbers. 
Figure~\ref{tab:table_example} gives an example table with one row.
Next, we generate a query $Q^{(i)}$ using predefined templates associated with its gold-standard answer $y^{(i)}$ on $\mathcal{T}^{(i)}$.
Our task consists of four types of natural language queries as summarized in Table~\ref{tab:query_example}, with annotated SQL-like logical forms for easy interpretation. 
We generate NL queries at various levels of compositionality to mimic the real world scenario.
The complexity of those queries ranges from simple \selwh/ queries to more complicated \nested/ ones involving multiple steps of computation.
Those queries are flexible enough to involve complex matching between NL phrases and logical constituents, which makes query understanding and execution nontrivial:
(1) the same field is described by different NL phrases (e.g., {\it ``How big is the country ...''} and {\it ``What is the size of the country ...''} for {\tt country\_size} field);
(2) different fields may be referred to by the same NL pattern (e.g, {\it ``\underline{in} China''} for {\tt country} and {\it ``\underline{in} Beijing''} for {\tt host\_city}); 
(3) simple NL constituents may be grounded to complex logical operations (e.g., {\it ``\underline{after} the game in Beijing''} implies comparing between {\tt year} fields).
In our experiments we use the above procedure to generate benchmark datasets consisting of different types of queries.
To make the artificial task harder, we enforce that all queries in the testing set do not appear in the training set.

\def\ds/{{\sc Mixtured}}
\def\dss/{{\sc Mixtured}-25K}
\def\dsl/{{\sc Mixtured}-100K}

To simulate the read-world scenario where queries of various types are issued to the model, we constructed two \ds/ datasets for our main experiments, with $25K$ and $100K$ training examples respectively, where four types of quires are sampled with the ratio $1:1:1:2$. Both  datasets share the same testing set of $20K$ examples, $5K$ for each type of query.

\begin{figure*}[t]
    \tabcolsep=0.04cm
    \centering
    \scriptsize
    \begin{tabular}{|c|c|c|c|c|c|c|c|c|c|}
    \hline
    {\bf year} & {\bf host\_city} & {\bf \#\_participants} & {\bf \#\_medals} & {\bf \#\_duration} & {\bf \#\_audience} & {\bf host\_country} & {\bf GDP} & {\bf country\_size} & {\bf population} \\
    \hline
    2008 & Beijing & 4,200 & 2,500 & 30 & 67,000 & China & 2,300 & 960 & 130 \\
    \hline
    \end{tabular}
    \caption{An example table in the synthetic QA task (only show one row)}
    \label{tab:table_example}
\end{figure*}


\begin{table*}[t]
    \centering
    \scriptsize
    \tabcolsep=0.05cm
    \begin{tabular}{l|l}
    \toprule
    \textbf{Query Type} & \textbf{Example Queries with Annotated SQL-like Logical Forms} \\
    \hline
        \multirow{4}{*}{\selwh/} 
                & $\triangleright$ $Q_1$: {\it How many people participated in the game in Beijing?}\\
                & ~~$F_1$: {\tt select \#\_participants, where host\_city = Beijing} \\
                & $\triangleright$ $Q_2$: {\it In which country was the game hosted in 2012?} \\
                & ~~$F_2$: {\tt select host\_country, where year = 2012} \\ \hline
        \multirow{4}{*}{\super/}
                & $\triangleright$ $Q_3$: {\it When was the lastest game hosted?} \\
                & ~~$F_3$: {\tt argmax(host\_city, year)} \\
                & $\triangleright$ $Q_4$: {\it How big is the country which hosted the shortest game?} \\
                & ~~$F_4$: {\tt argmin(country\_size, \#\_duration)} \\ \hline
        \multirow{4}{*}{\whsuper/} 
                & $\triangleright$ $Q_5$: {\it How long is the game with the most medals that has fewer than 3,000 participants?} \\
                & ~~$F_5$: {\tt where \#\_participants $<$ 3,000, argmax(\#\_duration, \#\_medals)} \\
                & $\triangleright$ $Q_6$: {\it How many medals are in the first game after 2008?} \\
                & ~~$F_6$: {\tt where \#\_year $>$ 2008, argmin(\#\_medals, \#\_year)} \\ \hline
        \multirow{5}{*}{\nested/}
                & $\triangleright$ $Q_7$: {\it Which country hosted the longest game before the game in Athens?} \\
                & ~~$F_7$: {\tt where year$<$(select year,where host\_city=Athens),argmax(host\_country,\#\_duration)} \\
                & $\triangleright$ $Q_8$: {\it How many people watched the earliest game that lasts for more days than the game in 1956?} \\
                & ~~$F_8$: {\tt where \#\_duration$<$(select \#\_duration,where year=1956),argmin(\#\_audience,\#\_year)} \\
    \bottomrule
    \end{tabular}
    \caption{Example queries in our synthetic QA task}
    \label{tab:query_example}
\end{table*}

\def\Qone/{$Q_5$}
\def\Qtwo/{$Q_7$}
\def\Qthree/{$Q_8$}
\def\FQone/{$F_5$}
\def\FQtwo/{$F_7$}
\def\FQthree/{$F_8$}


\subsection{Setup}

\noindent \textbf{[Tuning]} We use a \model/ with five executors. The number of layers for ${\sf DNN}_1^{(\ell)}$ and ${\sf DNN}_2^{(\ell)}$ are set to 2 and 3, respectively.
We set the dimensionality of word/entity embeddings and row/table annotations to 20, hidden layers to 50, and the hidden states of the GRU in query encoder to 150. $\alpha$ in Eq. \eqref{eq:sbs_objective} is set to 0.2. We pad the beginning of all input queries to a fixed size.

\model/ is trained via standard back-propagation. Objective functions are optimized using SGD in a mini-batch of size 100 with adaptive learning rates ({\sc AdaDelta}~\cite{DBLP:journals/corr/abs-1212-5701}). 
The model converges fast within 100 epochs. 

\noindent \textbf{[Baseline]} We compare our model with \sempre/~\cite{pasupat2015compositional}, a state-of-the-art semantic parser.

\noindent \textbf{[Metric]} We evaluate the performance of \model/ and \sempre/ (baseline) in terms of accuracy, defined as the fraction of correctly answered queries.

\subsection{Main Results}
\label{ssec:exp_results}


\begin{table*}[t]
    \centering
    \scriptsize
    \begin{tabular}{@{\extracolsep{4pt}}lccccccc}
    \toprule
    & \multicolumn{4}{c}{\sc{Mixtured-25K}} & \multicolumn{3}{c}{\sc{Mixtured-100K}} \\ \cline{2-5} \cline{6-8}
    & {\sc Sempre} & N2N & SbS & N2N - OOV & N2N & SbS & N2N - OOV \\ \hline
    \selwh/ & 93.8\% & 96.2\% & 99.7\% & 90.3\% & 99.3\% & 100.0\% & 97.6\% \\
    \super/ & 97.8\% & 98.9\% & 99.5\% & 98.2\% &  99.9\% & 100.0\% & 99.7\% \\
    \whsuper/ & 34.8\% & 80.4\% & 94.3\% & 79.1\% & 98.5\% & 99.8\% & 98.0\% \\
    \nested/ & 34.4\% & 60.5\% & 92.1\% & 57.7\% & 64.7\% & 99.7\% & 63.9\% \\ \hline
    Overall Acc. & 65.2\% & 84.0\% & 96.4\% & 81.3\% & 90.6\% & 99.9\% & 89.8\% \\
    \bottomrule
    \end{tabular}
    \caption{Accuracies on \ds/ datasets}
    \label{tab:acc_mixtured_dataset}
\end{table*}

Table~\ref{tab:acc_mixtured_dataset} summarizes the results of \sempre/ (baseline) and our \model/ under {\em end-to-end} (N2N) and {\em step-by-step} (SbS) settings. We show both the individual performance for each type of query and the overall accuracy. We evaluate \sempre/ only on \dss/ because of its long training time even on the smaller \dss/ ($>$ 3 days).
We give discussion of efficiency issues in Appendix~\ref{app:efficiency}.

We first discuss \model/'s performance under end-to-end (N2N) training setting (the 3rd and 6th column in Table~\ref{tab:acc_mixtured_dataset}), and defer the discussion for SbS setting to Section~\ref{ssec:sbs_training}. 
On \dss/, our model outperforms \sempre/ on all types of queries, with a marginal gain on {\em simple} queries (\selwh/, \super/), and significant improvement on {\em complex} ones (\whsuper/, \nested/).
When the size of training set grows (\dsl/), \model/ achieves near 100\% accuracy for the first three types of queries, while registering a decent overall accuracy of 90.6\%. 
These results suggest that our model is very effective in answering compositional natural language queries, especially those with complex semantic structures compared with the state-of-the-art system.

\begin{figure*}[t]
    \minipage{\textwidth}
    \centering
    \scriptsize
    \mbox{\Qone/: {\it How long is the game with the most medals that has fewer than 3,000 participants?}}
    \endminipage \vspace{5pt} \\
    \includegraphics[width=\textwidth]{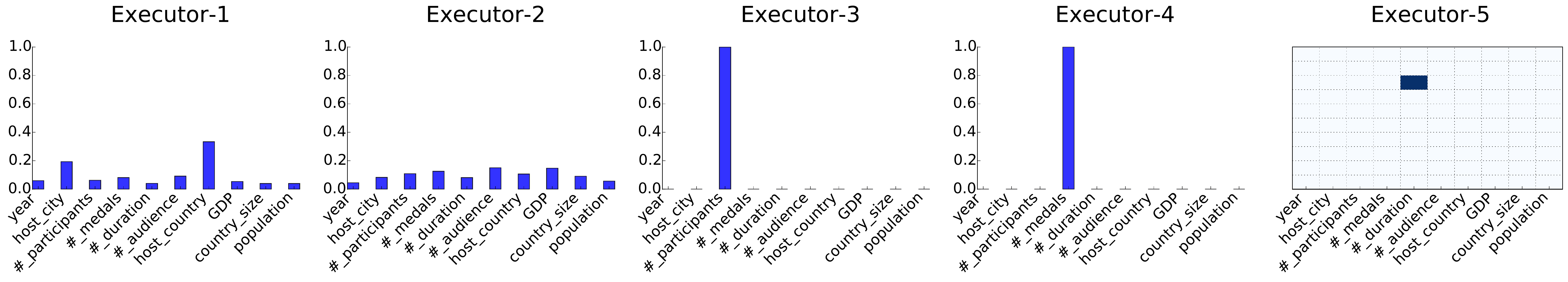}
    \caption{Weights visualization of query \Qone/}
    \label{fig:att_weight_q1}
\end{figure*}

\begin{figure*}[t]
    \captionsetup{skip=3pt}
    \minipage{\textwidth}
    \centering
    \scriptsize
    \mbox{\Qtwo/: {\it Which country hosted the longest game before the game in Athens?}}
    \endminipage \vspace{5pt} \\
    \includegraphics[width=\textwidth]{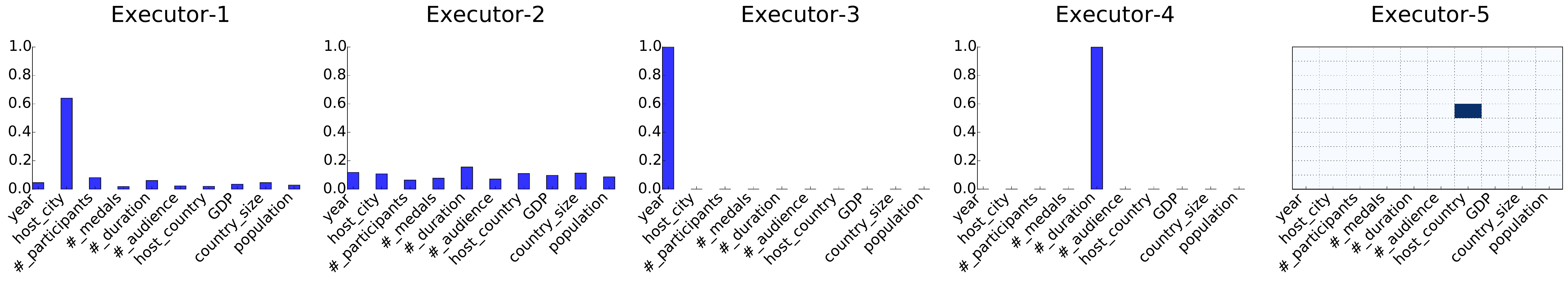}
    \caption{Weights visualization of query \Qtwo/}
    \label{fig:att_weight_q2}
\end{figure*}



\begin{figure*}[t]
    \captionsetup{skip=3pt}
    \minipage{\textwidth}
    \centering
    \scriptsize
    \mbox{\Qthree/: {\it How many people watched the earliest game that lasts for more days than the game in 1956?}}
    \endminipage \vspace{5pt} \\
    \includegraphics[width=\textwidth]{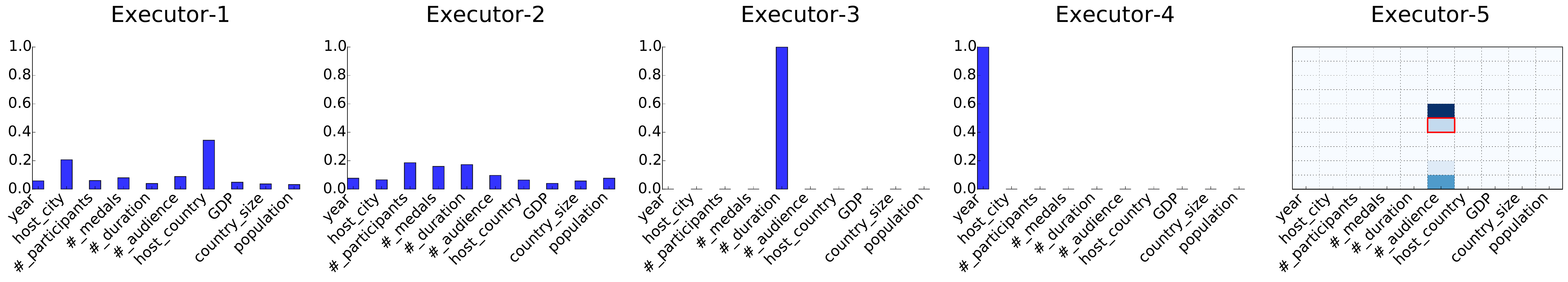}
    \caption{Weights visualization of query \Qthree/ (an incorrectly answered query)}
    \label{fig:att_weight_nested_wrong}
\end{figure*}

To further understand why our model is capable of handling compositional queries, we study the attention weights $\tilde{w}(\cdot)$ of Readers (Eq.~\ref{eq:read_attention_weight}) for executors in intermediate layers, and the answer probability (Eq.~\ref{eq:answer_prob}) the last executor outputs for each entry in the table. Those statistics are obtained from the model trained on \ds/-100K.
We sampled two queries (\Qone/ and \Qtwo/ in Table~\ref{tab:query_example}) in the dataset that our model answers correctly and visualized their corresponding values, as illustrated in Figure~\ref{fig:att_weight_q1} and~\ref{fig:att_weight_q2}, respectively.
We find that each executor actually {\em learns} its execution logic from just the correct answers in end-to-end training, which corresponds with our assumption.
For \Qone/, the model executes the query in three steps, with each of the last three executors performs a specific type of operation. For each row, {\sf Executor}-3 takes the value of the {\tt \#\_participants} field as input and computes intermediate annotations, while {\sf Executor}-4 focuses on the {\tt \#\_medals} field.
Finally, the last executor outputs high probability for the {\tt \#\_duration} field (the 5-th column) in the 3-rd row.
The attention weights for \textsf{Executor}-1 and \textsf{Executor}-2 appear to be meaningless because \Qone/ requires only three steps of execution, and the model learns to defer the meaningful execution to the last three executors. 
We can guess confidently that in executing \Qone/, {\sf Executor}-3 performs the conditional filtering operation ({\tt where} clause in \FQone/), and {\sf Executor}-4 performs the first part of {\tt argmax} (finding the maximum value of {\tt \#\_medals}), while the last executor finishes the execution by assigning high probability for the {\tt \#\_duration} field of the row with the maximum value of {\tt \#\_medals}.

Compared with the relatively simple \Qone/, \Qtwo/ is more complicated, whose logical form \FQtwo/ involves a nest sub-query, and requires five steps of execution. From the weights visualized in figure~\ref{fig:att_weight_q2}, we can find that the last three executors function similarly as the case in answering \Qone/, yet the execution logic for the first two executors is a bit obscure. 
We posit that this is because during end-to-end training, the supervision signal propagated from the top layer has decayed along the long path down to the first two executors, which causes vanishing gradient problem. 

We also investigate the case where our model fails to deliver the correct answer for complicated queries. 
Figure~\ref{fig:att_weight_nested_wrong} gives such a query \Qthree/ in Table~\ref{tab:query_example} together with its visualized weights. Similar as \Qtwo/, \Qthree/ requires five steps of execution. Besides messing up the weights in the first two executors, the last executor, {\sf Executor}-5, predicts a wrong entry as the query answer, instead of the highlighted (in red rectangle) correct entry.

\subsection{With Additional Step-by-Step Supervision}
\label{ssec:sbs_training}

\begin{figure*}[t]
    \vspace{10pt}
    \minipage{\textwidth}
    \centering
    \scriptsize
    \mbox{\Qtwo/: {\it Which country hosted the longest game before the game in Athens?}}
    \endminipage \vspace{5pt} \\
    \includegraphics[width=\textwidth]{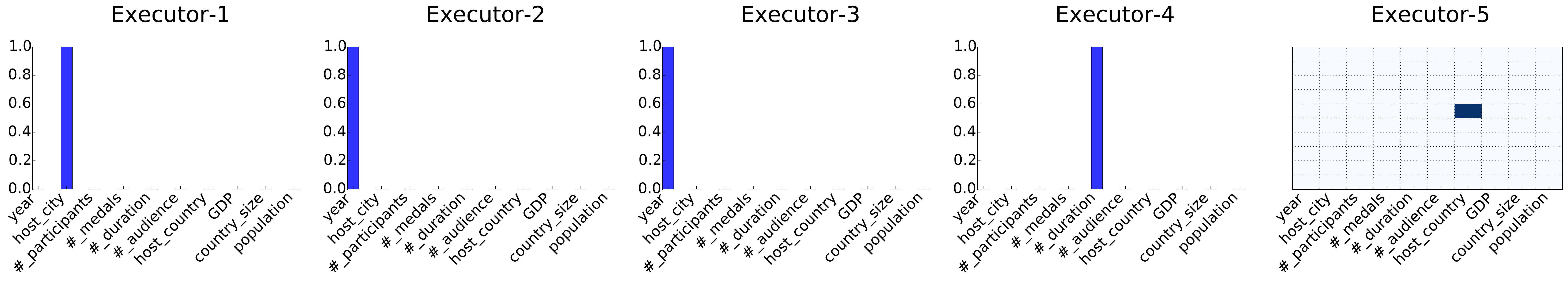}
    \caption{Weights visualization of query \Qtwo/ in step-by-step training setting}
    \label{fig:att_weight_q2_sbs}
\end{figure*}

To alleviate the vanishing gradient problem when training on complex queries as described in Section~\ref{ssec:exp_results}, in our next set of experiments we trained our \model/ model using step-by-step (SbS) training (Eq.~\ref{eq:sbs_objective}), where we encourage each executor to attend to a specific field that is known {\it a priori} to be relevant to its execution logic.
The results are shown in the 4-rd and 7-th columns of Table~\ref{tab:acc_mixtured_dataset}.
With stronger supervision signal, the model significantly outperforms the results in end-to-end setting, and achieves near $100\%$ accuracy on all types of queries, which shows that our proposed \model/ is capable of leveraging the additional supervision signal given to intermediate layers in SbS training setting, and answering complex and compositional queries with perfect accuracy.

Let us revisit the query \Qtwo/ in SbS setting with the weights visualization in Figure~\ref{fig:att_weight_q2_sbs}. In contrast to the result in N2N setting (Figure~\ref{fig:att_weight_q2}) where the attention weights for the first two executors are obscure, the weights in every executor are perfectly skewed towards the actual field pertain to each layer of execution (with a weight $1.0$). Quite interestingly, the attention weights for \textsf{Executor}-3 and \textsf{Executor}-4 are exactly the same with the result in N2N setting, while the weights for \textsf{Executor}-1 and \textsf{Executor}-2 are significantly different, suggesting \model/ learned a different execution logic in the SbS setting.

\subsection{Dealing with Out-Of-Vocabulary Words}
\label{ssec:oov_result}

\begin{figure*}[th!]
    \captionsetup{skip=3pt}
    \minipage{\textwidth}
    \centering
    \scriptsize
    \mbox{$Q_9$: {\it How many people watched the game in Macau?}}
    \endminipage \vspace{5pt} \\
    \includegraphics[width=\textwidth]{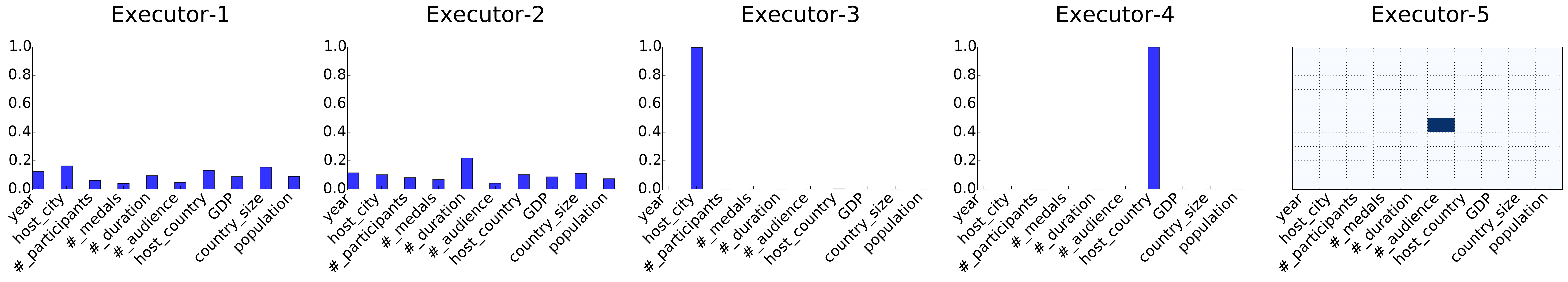}
    \caption{Weights visualization of query $Q_9$}
    \label{fig:att_weight_oov}
\end{figure*}

One of the major challenges for applying neural network models to NLP applications is to deal with Out-Of-Vocabulary (OOV) words, which is particularly severe for QA. It is hard to cover existing tail entities, while at the same time new entities appear in user-issued queries and back-end KBs everyday. Quite interestingly, we find that a simple variation of \model/ is able to handle unseen entities almost without any loss of accuracy.

Basically, we divide words in the vocabulary into {\em entity} words and {\em operation} words. Embeddings of entity words (e.g., {\it Beijing}, {\it China}) function in a way as index to facilitate the matching between entities in queries and tables during the layer-by-layer execution, and do not need to be updated once initialized;
while those of operation words, i.e., all non-entity words (e.g., numbers, {\it longest}, {\it before}, etc) carry semantic meanings relevant to execution and should be optimized during training. 
Therefore, after randomly initializing the embedding matrix $\mathbf{L}$, we only update the embeddings of operation words in training, while keeping those of entity words unchanged.

To test the model's performance with OOV words, we modify queries in the testing portion of the \ds/ dataset to replace all entity words (i.e., all country and city names) with OOV ones\footnote{They also have embeddings in $\mathbf{L}$.} unseen in the training set.
Results obtained using N2N training are summarized in the 5th and 8th columns of Table~\ref{tab:acc_mixtured_dataset}.
As it shows \model/ training in this OOV setting yields performance comparable to that in the non-OOV setting,
indicating that operation words and entity words play different roles in query execution.

An interesting question to investigate in this OOV setting is how \model/ distinguishes between different types of entity words (i.e., cities and countries) in queries, since their embeddings are randomly initialized and fixed thereafter. An example query is $Q_9:$ {\it ``How many people watched the game in \underline{Macau}?''}, where {\it Macau} is an OOV entity. To help understand how the model knows {\it Macau} is a city, we give its weights visualization in Figure~\ref{fig:att_weight_oov}. Interestingly, the model first checks the {\tt host\_city} field in {\sf Executor}-3, and then {\tt host\_country} in {\sf Executor}-4, which suggests that the model learns to scan all possible fields where the OOV entity may belong to. 

\subsection{Simulating Large Knowledge Source}

\begin{table}[tb]
    \centering
    \begin{tabular}{l|ccc|c}
    \toprule
    Query Type & \selwh/ & \super/ & \whsuper/ & Overall \\ \hline
    Accuracy & 68.2\% & 84.8\% & 80.2\% & 77.7\% \\
    \bottomrule
    \end{tabular}
    \caption{Accuracies for large knowledge source simulation}
    \label{tab:large_kb_src}
\end{table}

\begin{figure}[tb]
    \centering
    \includegraphics[scale=0.6]{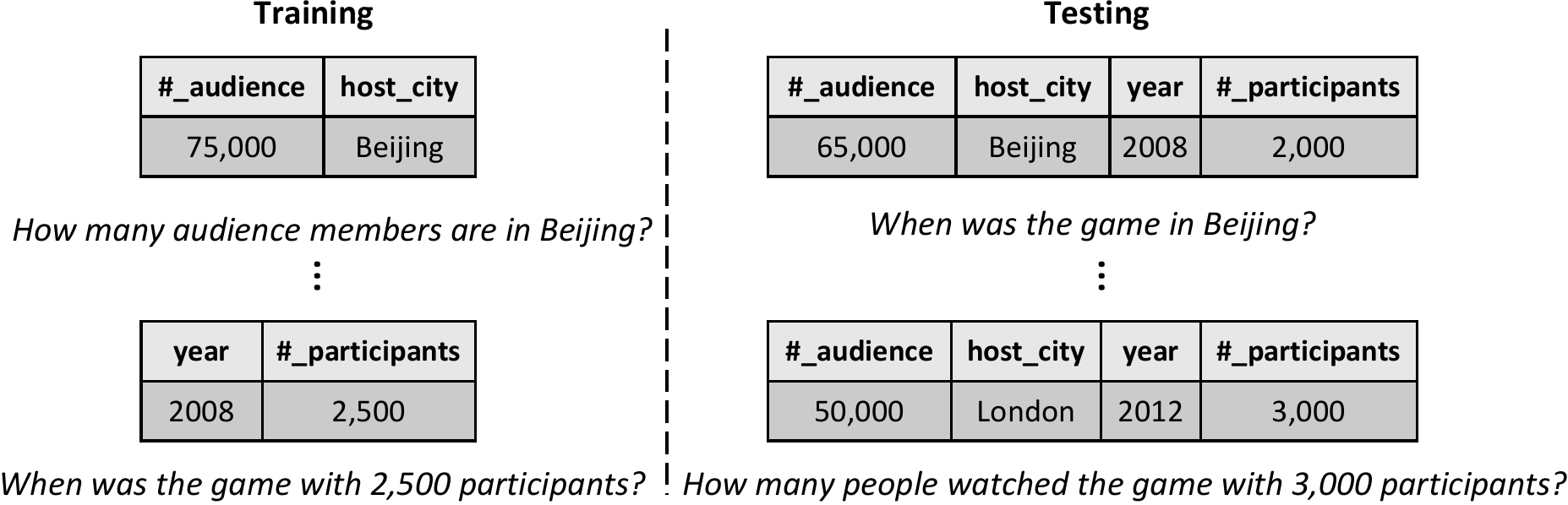}
    \caption{Large knowledge source simulation}
    \label{fig:large_kb_src}
\end{figure}

An important direction in semantic parsing research is to scale to large knowledge source~\cite{berant2013freebase,pasupat2015compositional}. In this set of experiments we simulate a test case to evaluate \model/'s ability to generalize to large knowledge source. We train a model on tables whose field sets are either $\mathcal{F}_1, \mathcal{F}_2, \ldots, \mathcal{F}_5$, where $\mathcal{F}_i$ (with $|\mathcal{F}_i|=5$) is a subset of the entire set $\mathcal{F}_{\mathcal{T}}$. We then test the model on tables with all fields $\mathcal{F}_{\mathcal{T}}$ and queries whose fields span multiple subsets $\mathcal{F}_i$. Figure~\ref{fig:large_kb_src} illustrates the setting. Note that all testing queries exhibit field combinations unseen in the training data, to mimic the difficulty the system often encounter when scaling to large knowledge source, which usually poses great challenge on model's generalization ability.
We then train and test the model only on a new dataset of the first three types of relatively simple queries (namely \selwh/, \super/ and \whsuper/). The sizes of training/testing splits are 75,000 and 30,000, with equal numbers for different query types. Table~\ref{tab:large_kb_src} lists the results. \model/ still maintains a reasonable performance even when the compositionality of testing queries is previously unseen, showing the model's generalization ability in tackling unseen query patterns through composition of familiar ones, and hence the potential to scale to larger and unseen knowledge sources.


\section{Related Work}

Our work falls into the research area of Semantic Parsing, where the key problem is to parse Natural Language queries into logical forms executable on KBs. Classical approaches for Semantic Parsing can be broadly divided into two categories.
The first line of research resorts to the power of grammatical formalisms (e.g., Combinatory Categorial Grammar) to parse NL queries and generate corresponding logical forms, which requires curated/learned lexicons defining the correspondence between NL phrases and symbolic constituents~\cite{DBLP:conf/uai/ZettlemoyerC05,DBLP:conf/emnlp/KwiatkowskiCAZ13,DBLP:conf/emnlp/ArtziLZ15,DBLP:conf/emnlp/ZettlemoyerC07}. The model is tuned with annotated logical forms, and is capable of recovering complex semantics from data, but often constrained on a specific domain due to scalability issues brought by the crisp grammars and the lack of annotated training data.
Another line of research takes a semi-supervised learning approach, and adopts the results of query execution (i.e., answers) as supervision signal~\cite{chen:icml08,berant2013freebase,pasupat2014extraction,pasupat2015compositional,Yih15Acl}.
The parsers, designed towards this new learning paradigm, take different types of forms, ranging from generic chart parsers~\cite{berant2013freebase,pasupat2015compositional} to more specifically engineered, task-oriented ones~\cite{Yih15Acl,DBLP:conf/acl/MisraTLS15}. Semantic parsers in this category often scale to open domain knowledge sources, but lack the ability of understanding compositional queries because of the intractable search space incurred by the flexibility of parsing algorithms. Our work follows this line of research in using query answers as indirect supervision to facilitate end-to-end training using QA tasks, but performs semantic parsing in distributional spaces, where logical forms are ``neuralized'' to an executable distributed representation.

Our work is also related to the recent advances of modeling symbolic computation using Deep Neural Networks. Pioneered by the development of Neural Turing Machines (NTMs)~\cite{DBLP:journals/corr/GravesWD14}, this line of research studies the problem of using differentiable neural networks to perform ``hard'' symbolic execution. 
As an independent line of research with similar flavor, Zaremba et al.~\cite{DBLP:journals/corr/ZarembaS14} designed a LSTM-RNN to execute simple Python programs, where the parameters are learned by comparing the neural network output and the correct answer.
Our work is related to both lines of work, in that like NTM, we heavily use external memory and flexible way of processing (e.g., the attention-based reading in the operations in \textsf{Reader}) and like ~\cite{DBLP:journals/corr/ZarembaS14}, \model/ learns to execute a sequence with complicated structure, and the model is tuned from the executing them. As a highlight and difference from the previous work, we have a deep architecture with multiple layer of external memory, with the neural network operations highly customized to querying KB tables.

Perhaps the most related work to date is the recently published {\sc Neural Programmer} proposed by Neelakantan et al.~\cite{2015arXiv151104834N}, which studies the same task of executing queries on tables with Deep Neural Networks. {\sc Neural Programmer} uses a neural network model to select operations during query processing. While the query planning (i.e., which operation to execute at each time step) phase is modeled softly using neural networks, the symbolic operations are predefined by users. In contrast \model/ is fully distributional:  it models both the query planning and the operations with neural networks, which are jointly optimized via end-to-end training. Our \model/ model learns symbolic operations using data-driven approach, and demonstrates that a fully neural, end-to-end differentiable system is capable of modeling and executing compositional arithmetic and logic operations upto certain level of complexity.


\section{Conclusion and Future Work}


In this paper we propose \model/, a fully neural, end-to-end differentiable network that learns to execute queries on tables. 
We present results on a set of synthetic QA tasks to demonstrate the ability of \model/ to answer fairly complicated compositional queries across multiple tables. 
In the future we plan to advance this work in the following directions.
First we will apply \model/ to natural language questions and natural language answers, where both the input query and the output supervision are noisier and less informative. 
Second, we are going to scale to real world QA task as in~\cite{pasupat2015compositional}, for which we have to deal with a large vocabulary and novel predicates. Third, we are going to work on the computational efficiency issue in query execution by heavily borrowing the symbolic operation.

\bibliographystyle{abbrv}
\small{\bibliography{neural_enquirer}}

\newpage
\appendix

\section{Computation of Query Encoder}
\label{app:gru}

We use a bidirectional RNN as the Query Encoder, which consists of a {\it forward} GRU and a {\it backward} GRU. Given the sequence of word embeddings of $Q$: $\{ \mathbf{x}_1, \mathbf{x}_2, \ldots, \mathbf{x}_T \}$,
at each time step $t$, the forward GRU computes the hidden state $\mathbf{h}_t$ as follows:
\begin{align*}
  \mathbf{h}_t &= \mathbf{z}_t \mathbf{h}_{t-1} + (\mathbf{1} - \mathbf{z}_t) \mathbf{\tilde{h}}_t \\
  \mathbf{\tilde{h}}_t &= \tanh ( \mathbf{W}\mathbf{x}_t + \mathbf{U}(\mathbf{r}_t \circ \mathbf{h}_{t-1} ) ) \\
  \mathbf{z}_t &= \sigma (\mathbf{W}_z\mathbf{x}_t + \mathbf{U}_z\mathbf{h}_{t-1}) \\
  \mathbf{r}_t &= \sigma (\mathbf{W}_r\mathbf{x}_t + \mathbf{U}_r\mathbf{h}_{t-1})
\end{align*}
where $\mathbf{W}$, $\mathbf{W}_z$, $\mathbf{W}_r$, $\mathbf{U}$, $\mathbf{U}_z$, $\mathbf{U}_r$ are parametric matrices, $\mathbf{1}$ the column vector of all ones, and $\circ$ element-wise multiplication. The backward GRU reads the sequence in reverse order. 
We concatenate the last hidden states given by the two GRUs as the vectorial representation $\q/$ of the query.

\section{Efficiency of Model Learning}
\label{app:efficiency}

\begin{figure}[h]
	\centering
	\includegraphics[scale=0.4]{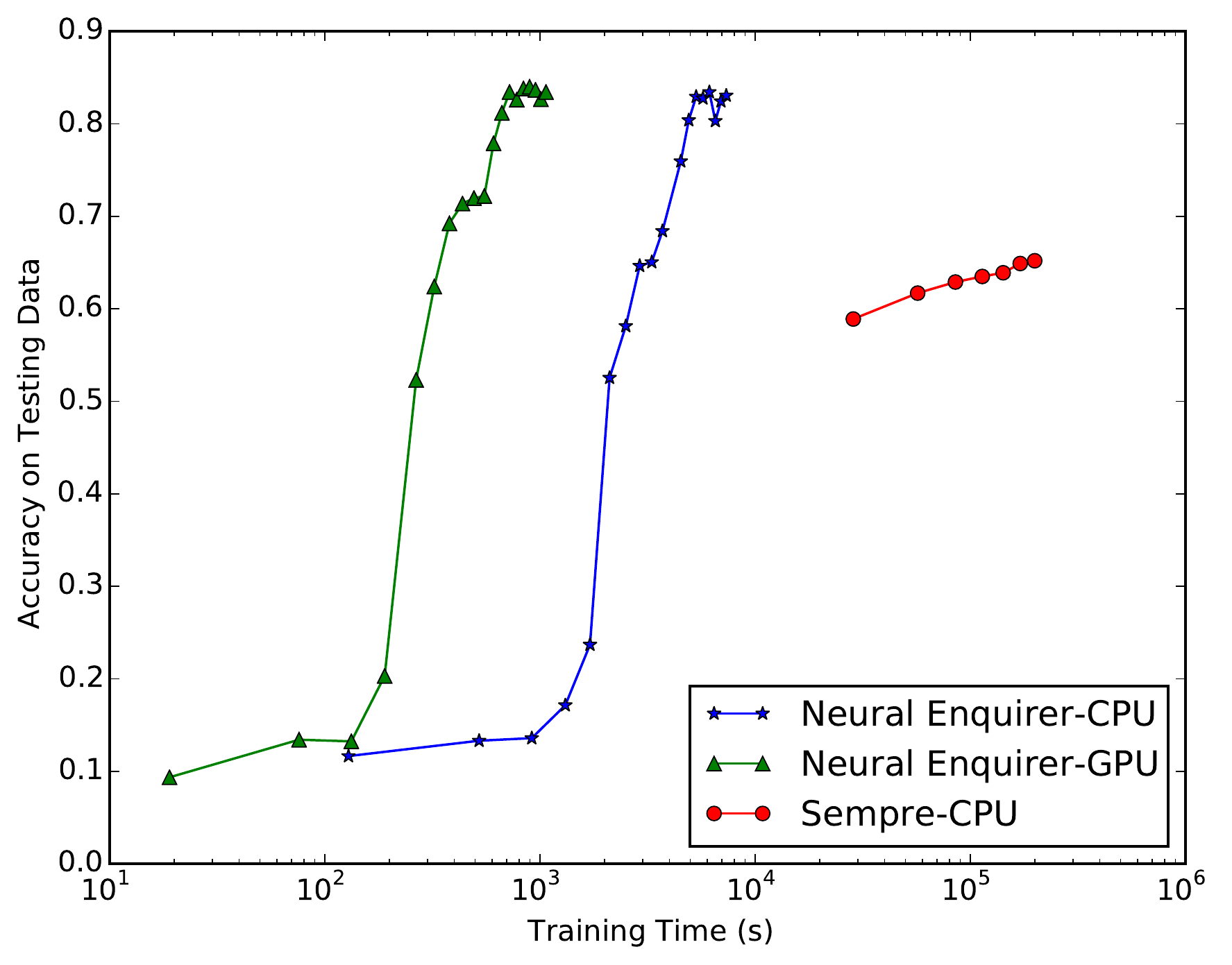}
	\caption{Accuracy on Testing by Training Time}
	\label{fig:acc_by_time}
\end{figure}

We compared the efficiency of \model/ and \sempre/ (baseline) in training by plotting the accuracy on testing data by training time. Figure~\ref{fig:acc_by_time} illustrates the results. We train \model/-CPU and \sempre/ on a machine with Intel Core i7-3770@3.40GHz and 16GB memory, while \model/-GPU is tuned on Nvidia Tesla K40. \model/-CPU is 10 times faster than \sempre/, and \model/-GPU is 100 times faster.

\section{Handling Multiple Tables}
\label{app:multi_table}

\subsection{\modelm/ Model}

\begin{figure}[tb]
	\centering
	\includegraphics[width=0.8\textwidth]{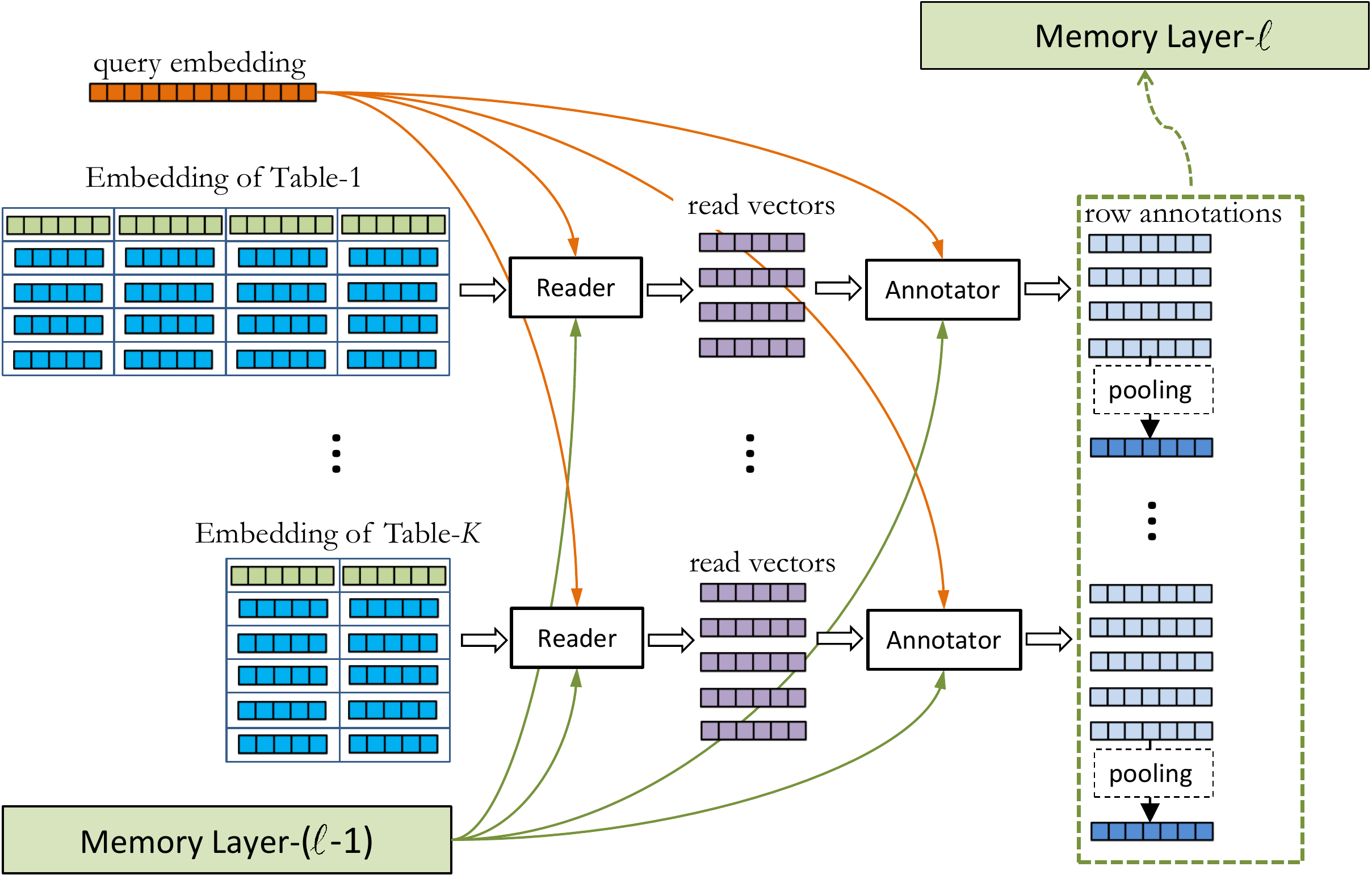}
	\caption{\textsf{Executor}-$(\ell, 1)$ and \textsf{Executor}-$(\ell, K)$ in multiple tables case}
	\label{fig:executor_multi_tables}
\end{figure}

\begin{figure}[tb]
	\centering
	\includegraphics[width=0.8\textwidth]{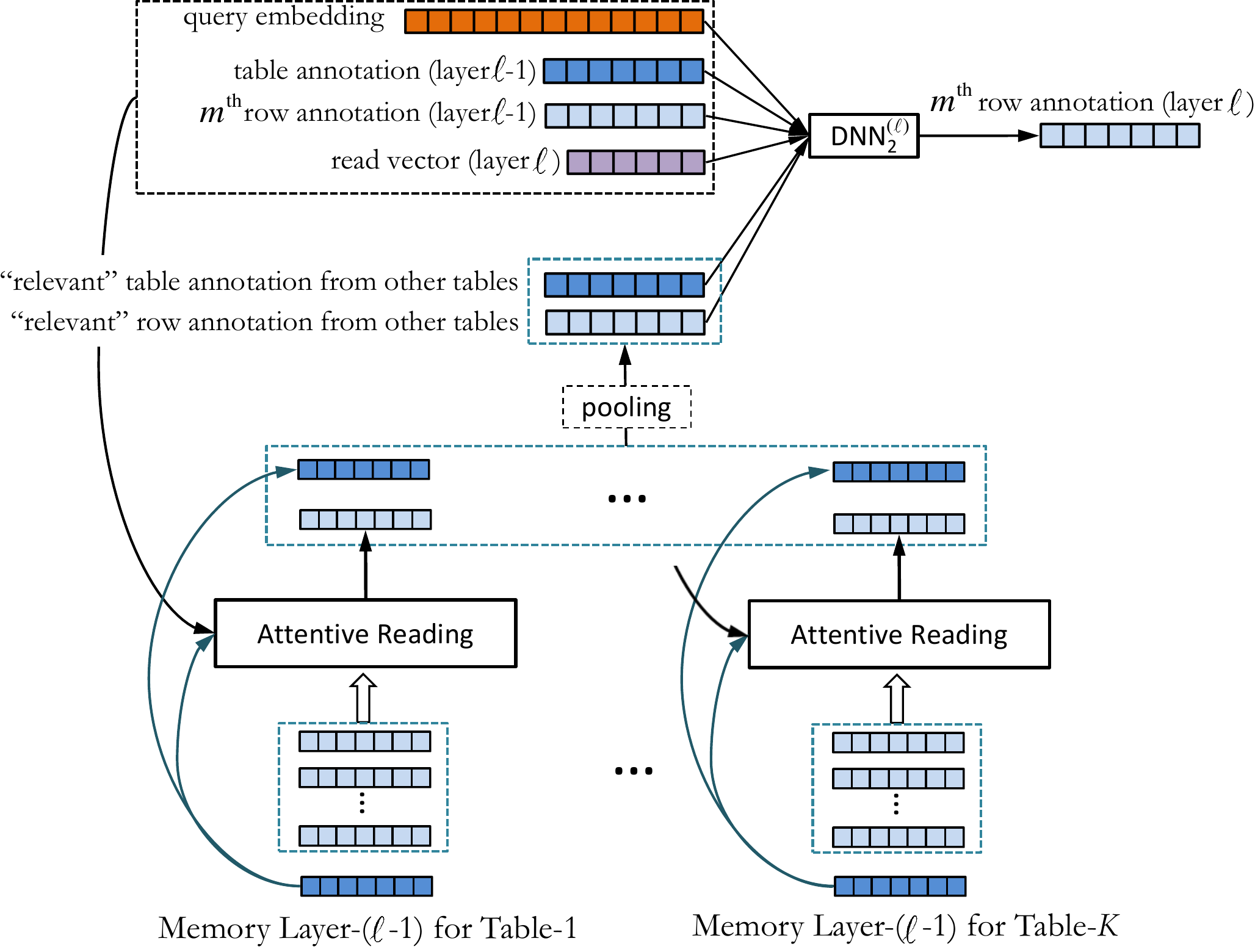}
	\caption{Illustration of Annotator for multiple tables case}
	\label{fig:annotator_multi_tables}
\end{figure}

Basically, \modelm/ assigns an executor to each table $\mathcal{T}_k$ in every execution layer $\ell$, denoted as {\sf Executor-$(\ell, k)$}.
Figure~\ref{fig:executor_multi_tables} pictorially illustrates \textsf{Executor}-$(\ell, 1)$ and \textsf{Executor}-$(\ell, K)$ working on \textsf{Table-1} and \textsf{Table-K} respectively. 
Within each executor, the Reader is designed the same way as single table case, while we modify the Annotator to let in the information from other tables. 
More specifically, for {\sf Executor-$(\ell, k)$}, we modify its Annotator by extending Eq.~\eqref{eq:row_annot} to leverage computational results from other tables when computing the annotation for the $m$-th row:
\begin{align*}
  \mathbf{a}^{\ell}_{k, m} &= f_\textsc{a}^{\ell}(\mathbf{r}^{\ell}_{k, m}, \q/, \mathbf{a}^{\ell-1}_{k, m}, \mathbf{g}^{\ell-1}_{k}, \hat{\mathbf{a}}^{\ell-1}_{k, m}, \hat{\mathbf{g}}^{\ell-1}_{k})
  \\ &= \textsf{DNN}^{(\ell)}_2([\mathbf{r}^{\ell}_{k, m}; \q/; \mathbf{a}^{\ell-1}_{k, m}; \mathbf{g}^{\ell-1}_{k}; \hat{\mathbf{a}}^{\ell-1}_{k, m}; \hat{\mathbf{g}}^{\ell-1}_{k}])
\end{align*}
This process is illustrated in Figure~\ref{fig:annotator_multi_tables}. Note that we add subscripts $k \in [1, K]$ to the notations to index tables. 
To model the interaction between tables, the Annotator incorporates the ``relevant'' row annotation, $\hat{\mathbf{a}}^{\ell-1}_{k, m}$, and the ``relevant'' table annotation, $\hat{\mathbf{g}}^{\ell-1}_{k}$ derived from the previous execution results of other tables when computing the current row annotation. 

A relevant row annotation stores the data fetched from row annotations of other tables, while a relevant table annotation summarizes the table-wise execution results from other tables. We now describe how to compute those annotations. First, for each table $\mathcal{T}_{k'}$, $k'\neq k$, we fetch a relevant row annotation $\bar{\mathbf{a}}^{\ell-1}_{k, k', m}$ from all row annotations $\{\mathbf{a}^{\ell-1}_{k', m'}\}$ of $\mathcal{T}_{k'}$ via attentive reading:
\[
  \bar{\mathbf{a}}^{\ell-1}_{k, k', m} = \sum_{m'=1}^{M_{k'}} \frac{\exp (\gamma(\mathbf{r}^{\ell}_{k, m}, \q/, \mathbf{a}^{\ell-1}_{k, m}, \mathbf{a}^{\ell-1}_{k', m'}, \mathbf{g}^{\ell-1}_{k}, \mathbf{g}^{\ell-1}_{k'}))}
  {\sum_{m''=1}^{M_{k'}} \exp (\gamma(\mathbf{r}^{\ell}_{k, m}, \q/, \mathbf{a}^{\ell-1}_{k, m}, \mathbf{a}^{\ell-1}_{k', m''}, \mathbf{g}^{\ell-1}_{k}, \mathbf{g}^{\ell-1}_{k'}))}
  \mathbf{a}^{\ell-1}_{k', m'}.
\]
Intuitively, the attention weight $\gamma(\cdot)$ (modeled by a DNN) captures how important the $m'$-th row annotation from table $\mathcal{T}_{k'}$, $\mathbf{a}^{\ell-1}_{k', m'}$ is with respect to the current step of execution.
After getting the set of row annotations fetched from all other tables, $\{ \bar{\mathbf{a}}^{\ell-1}_{k, k', m} \}_{k'=1, k' \neq k}^{K}$, we then compute $\hat{\mathbf{a}}^{\ell-1}_{k, m}$ and $\hat{\mathbf{g}}^{\ell-1}_{k}$ via a pooling operation\footnote{This operation is trivial in our experiments on two tables.} on $\{ \bar{\mathbf{a}}^{\ell-1}_{k, k', m} \}_{k'=1, k' \neq k}^{K}$ and $\{ \mathbf{g}^{\ell-1}_{k'} \}_{k'=1, k' \neq k}^{K}$:
\[
  \langle\hat{\mathbf{a}}^{\ell-1}_{k, m}, \hat{\mathbf{g}}^{\ell-1}_{k}\rangle = 
  \hat{f}_{\textsc{Pool}}(\{\bar{\mathbf{a}}^{\ell-1}_{k, 1, m}, \bar{\mathbf{a}}^{\ell-1}_{k, 2, m}, \ldots, \bar{\mathbf{a}}^{\ell-1}_{k, K, m}\}; \{\mathbf{g}^{\ell-1}_{1}, \mathbf{g}^{\ell-1}_{2}, \ldots, \mathbf{g}^{\ell-1}_{K}\}).
\]
In summary, relevant row and table annotations encode the local and global computational results from other tables. By incorporating them into calculating row annotations, \modelm/ is capable of answering queries that involve interaction between multiple tables.

Finally, \modelm/ outputs the ranked list of answer probabilities by normalizing the value of $g(\cdot)$ in Eq. \eqref{eq:answer_prob} over each entry for very table.

\subsection{Experimental Results}

\label{sec:exp_multi_table}

\begin{figure}[t]
    \captionsetup{skip=-1pt}
    \centering
    \small
    \caption*{\sf Table-1}
    \begin{tabular}{|c|c|c|c|c|c|c|}
    \hline
    {\bf country} & {\bf year} & {\bf host\_city} & {\bf \#\_participants} & {\bf \#\_medals} & {\bf \#\_duration} & {\bf \#\_audience} \\
    \hline
    China & 2008 & Beijing & 3,500 & 4,200 & 30 & 67,000 \\
    \hline
    \end{tabular}
    \caption*{\sf Table-2}
    \begin{tabular}{|c|c|c|c|}
    \hline
    {\bf country} & {\bf continent} & {\bf population} & {\bf country\_size} \\
    \hline
    China & Asia & 130 & 960 \\
    \hline
    \end{tabular}
    \vspace{4pt}
    \caption{Multiple tables example in the synthetic QA task (only show one row)}
    \label{tab:table_example_multi}
\end{figure}

We present preliminarily results for \modelm/, which we evaluated on SQL-like \selwh/ logical forms (like $F_1$, $F_2$ in Table~\ref{tab:query_example}). We sampled a dataset of $100K$ examples, with each example having two tables as in Figure~\ref{tab:table_example_multi}. Out of all \selwh/ queries, roughly half of the queries (denoted as ``Join'') require joining the two tables to derive answers. We tested on a model with three executors. Table~\ref{tab:acc_multi_table} lists the results. The accuracy of join queries is lower than that of non-join queries, which is caused by additional interaction between the two tables involved in answering join queries.

\begin{table}[h!]
    \centering
    \begin{tabular}{lcccc}
    \toprule
    \textbf{Query Type} & Non-Join & Join & Overall \\
    \midrule
    \textbf{Accuracy} & $99.7\%$ & $81.5\%$ & $91.3\%$ \\
    \bottomrule
    \end{tabular}
    \caption{Accuracies of \selwh/ queries on two tables}
    \label{tab:acc_multi_table}
\end{table}

\begin{figure*}[h!]
    \minipage{\textwidth}
    \centering
    \mbox{$Q_9$: {\tt select country\_size, where year = 2012}}
    \endminipage \vspace{2pt} \\
    \minipage{0.166\textwidth}
    \includegraphics[width=\textwidth]{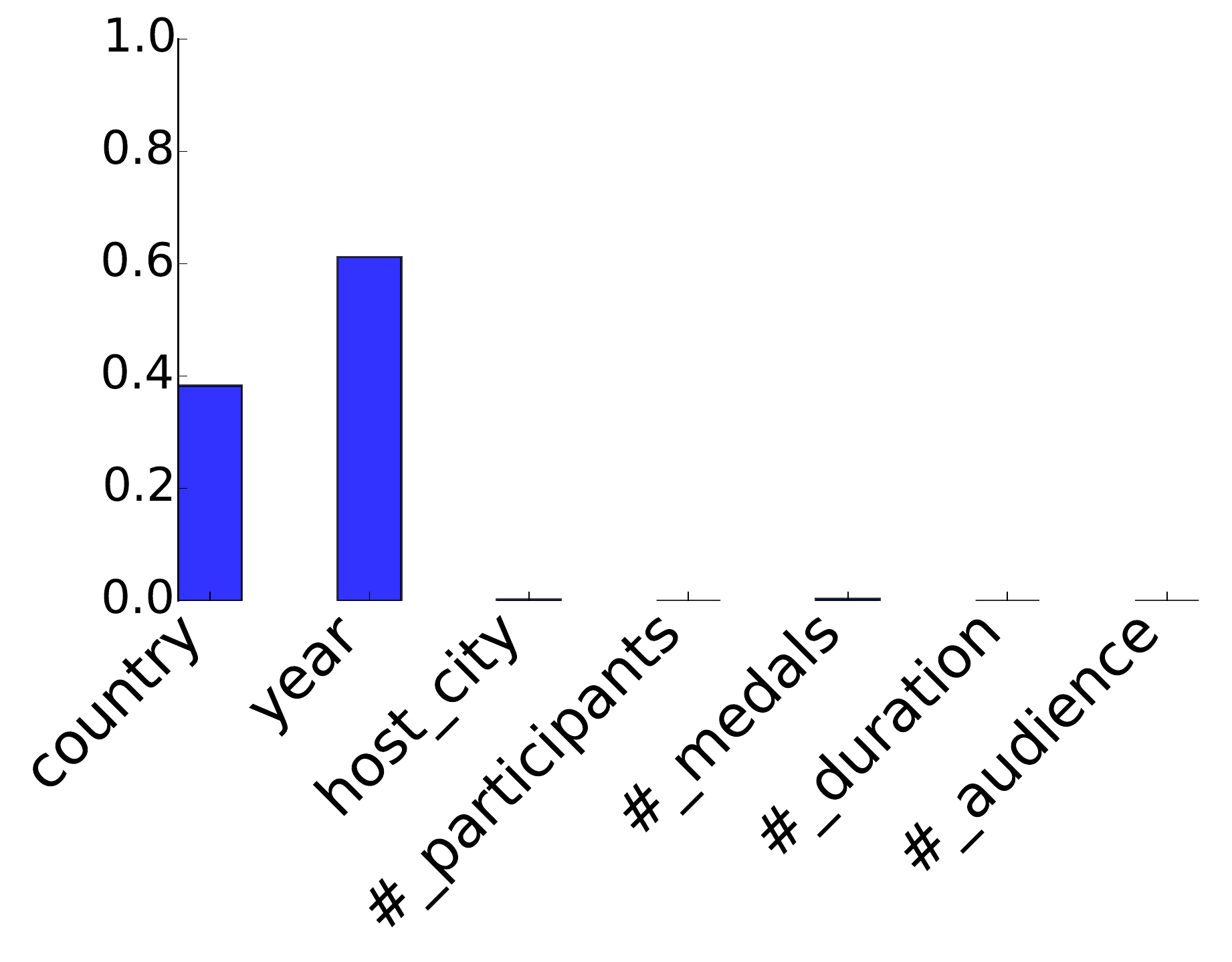}
    \captionsetup{font=small,skip=2pt,justification=centering}
    \caption*{{\sf Executor-$(1, 1)$}}
    \endminipage\hfill
    \minipage{0.166\textwidth}
    \centering
    \includegraphics[height=2cm]{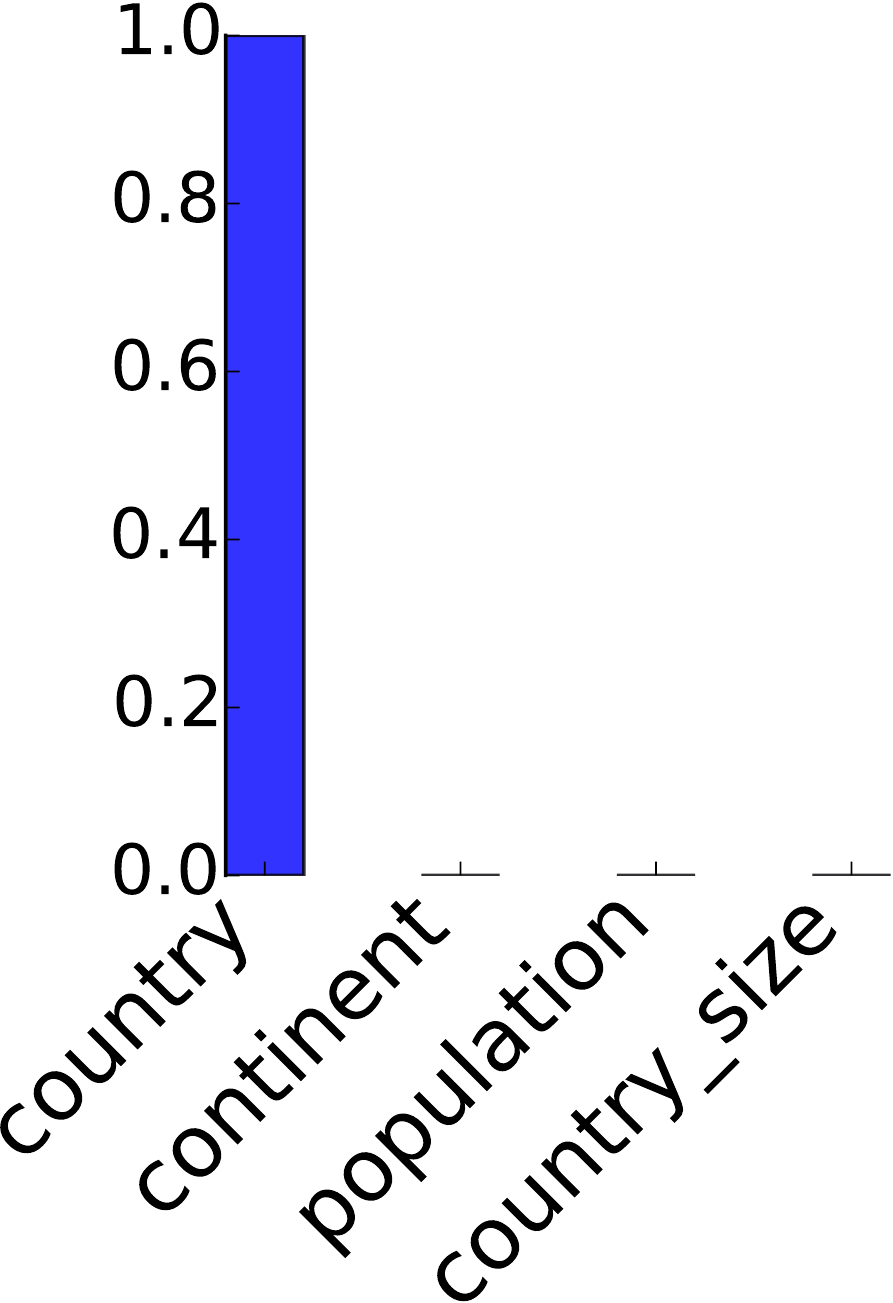}
    \captionsetup{font=small,skip=2pt,justification=centering}
    \caption*{{\sf Executor-$(1, 2)$}}
    \endminipage\hfill
    \minipage{0.166\textwidth}%
    \includegraphics[width=\textwidth]{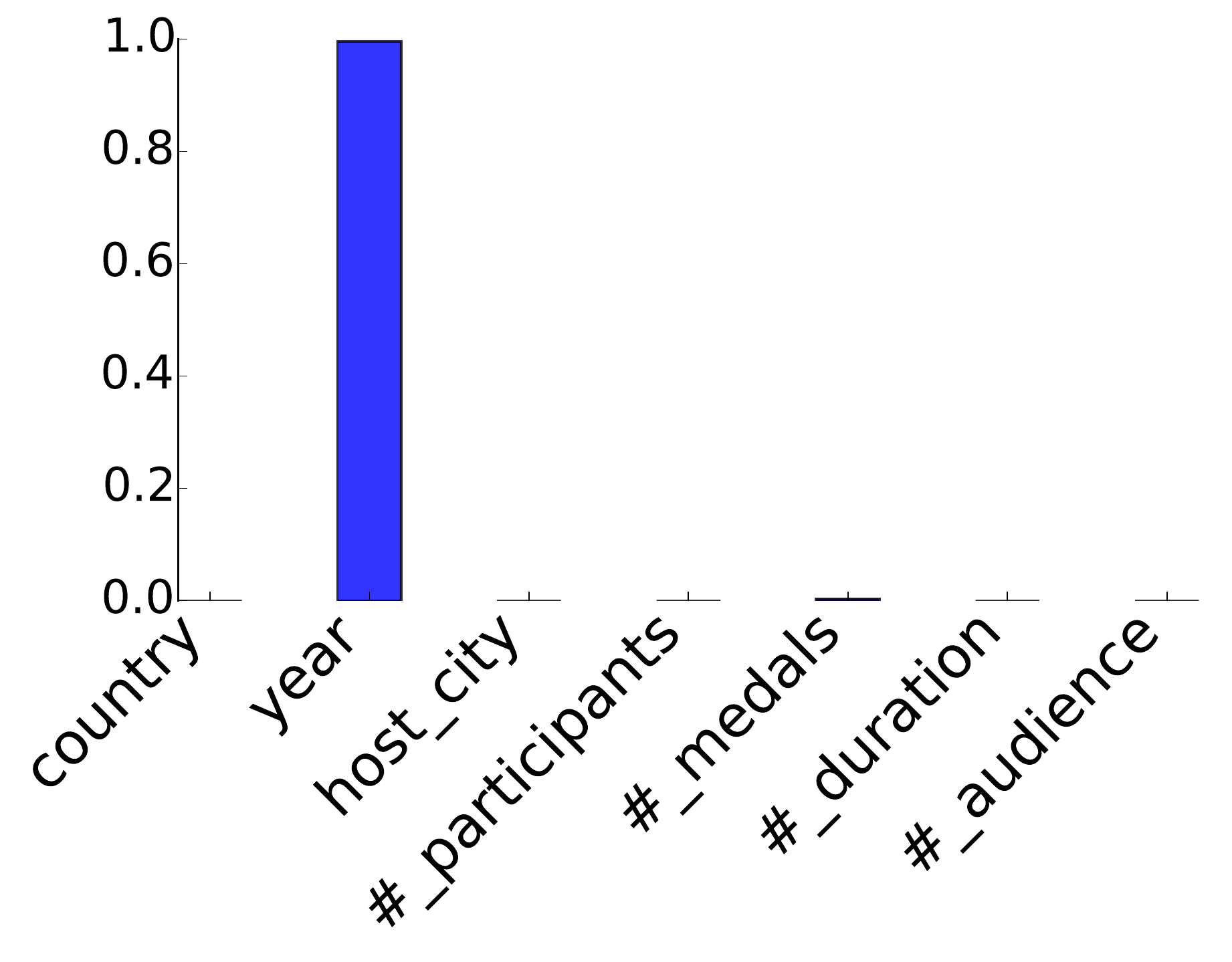}
    \captionsetup{font=small,skip=2pt,justification=centering}
    \caption*{{\sf Executor-$(2, 1)$}}
    \endminipage\hfill
    \minipage{0.166\textwidth}%
    \centering
    \includegraphics[height=2cm]{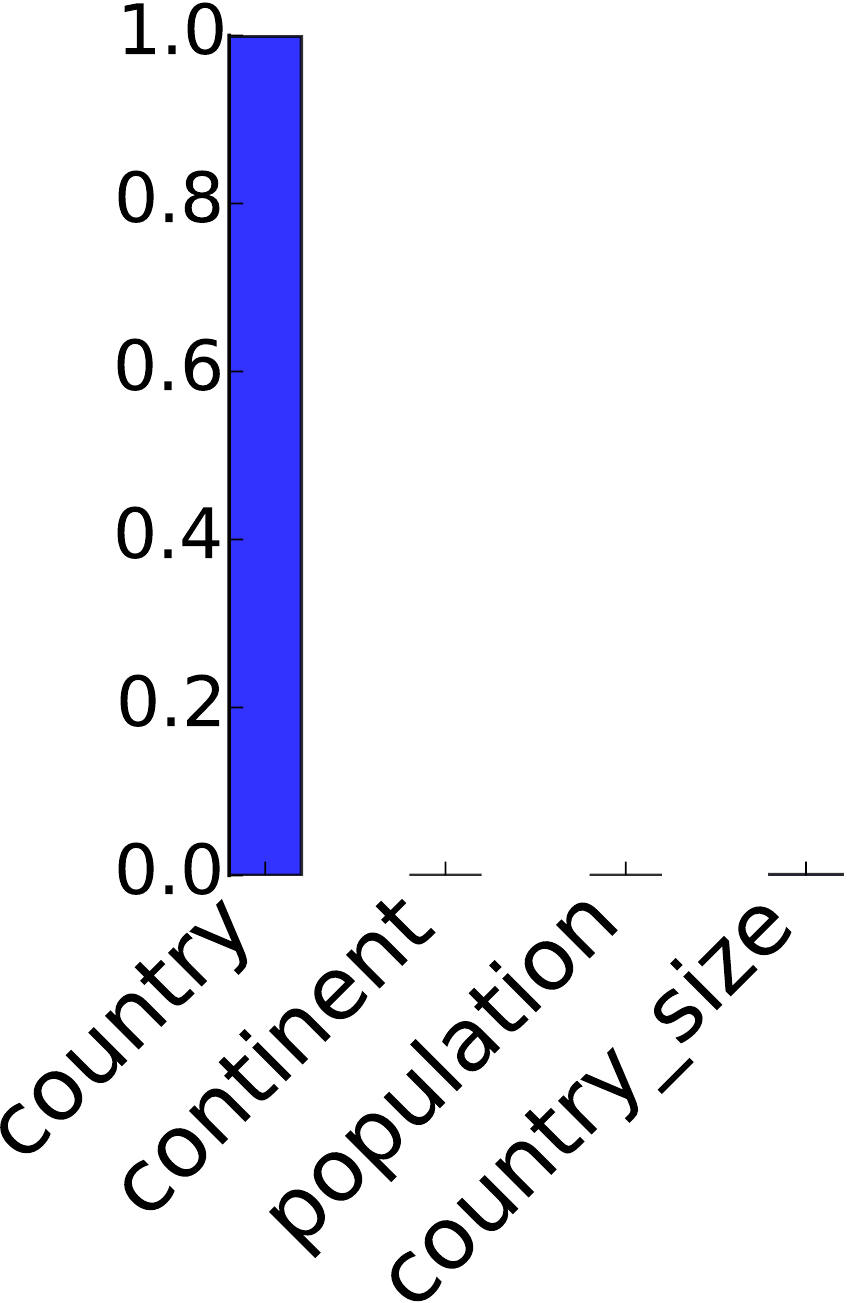}
    \captionsetup{font=small,skip=2pt,justification=centering}
    \caption*{{\sf Executor-$(2, 2)$}}
    \endminipage\hfill
    \minipage{0.14\textwidth}%
    \centering
    \includegraphics[height=2.1cm]{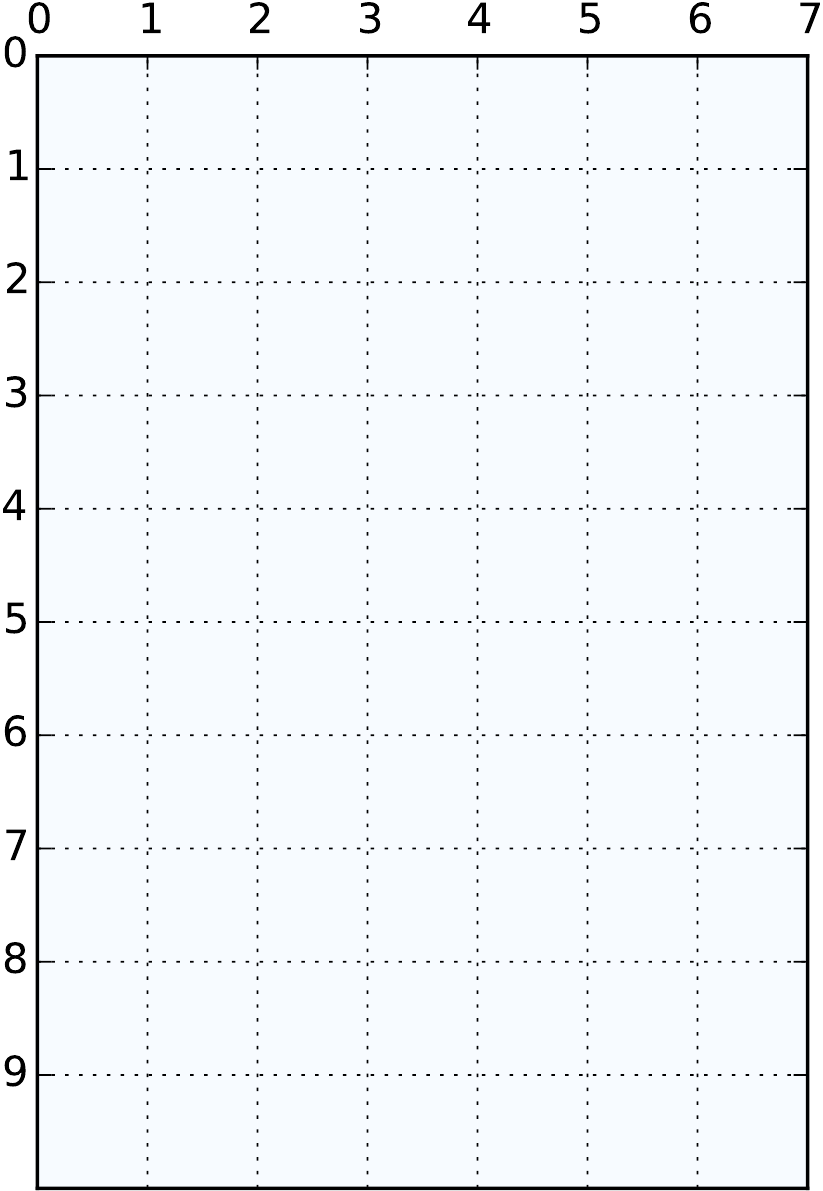}
    \captionsetup{font=small,skip=0pt,justification=centering}
    \caption*{{\sf Executor-$(3, 1)$}}
    \endminipage\hfill
    \minipage{0.16\textwidth}%
    \centering
    \includegraphics[height=2.1cm]{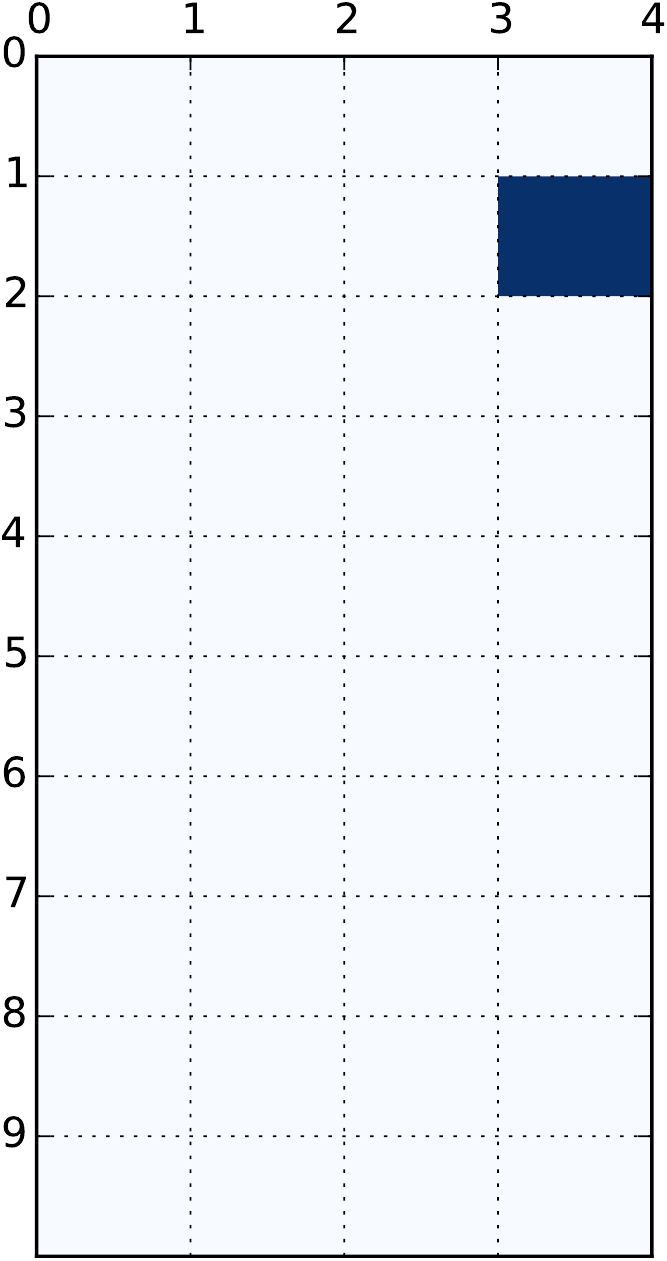}
    \captionsetup{font=small,skip=0pt,justification=centering}
    \caption*{{\sf Executor-$(3, 2)$}}
    \endminipage

    \caption{Weights visualization of query $Q_9$}
    \label{fig:att_weight_q4}
\end{figure*}

We find that \modelm/ is capable of identifying that the {\it country} field is the foreign key linking the two tables. Figure~\ref{fig:att_weight_q4} illustrates the attention weights for a correctly answered join query $Q_9$. Although the query does not contain any hints for the foreign key ({\it country} field), {\sf Executor-$(1,1)$} (the executor at Layer-1 on {\sf Table-1}) operates on an ensemble of embeddings of the {\it country} and {\it year} fields, whose outputting row annotations (contain information of both the key {\it country} and the value {\it year}) are sent to {\sf Executor-$(2,2)$} to compare with the {\it country} field in {\sf Table-2}.
We posit that the result of comparison is stored in the row annotations of {\sf Executor-$(2,2)$} and subsequently sent to the executors at Layer-3 for computing the answer probability for each entry in the two tables.

\end{document}